  \documentclass[final,5p,times,twocolumn]{elsarticle}

\usepackage{amssymb}
\usepackage{amsmath}
\usepackage{graphicx}
\usepackage{booktabs}
\usepackage{tabularx} 
\usepackage{afterpage}
\usepackage{multirow}
\usepackage[dvipsnames]{xcolor}
\usepackage{cuted}  
\usepackage[small, tableposition=top]{caption} 
\usepackage{algorithm}
\usepackage{algpseudocode}
\usepackage{hyperref}

\journal{.}

\begin{document}

\begin{frontmatter}



\title{Bayes-CATSI: A variational Bayesian deep learning framework for medical time series data imputation}


\author[inst1,inst3]{Omkar Kulkarni}
\author[inst2,inst3]{Rohitash Chandra}

\affiliation[inst2]{Transitional Artificial Intelligence Research Group, School of Mathematics and Statistics, UNSW Sydney, Australia}

\affiliation[inst3]{Centre for  Artificial Intelligence and Innovation, Pingala Institute, Sydney, Australia}

\affiliation[inst1]{organization={Department of Economics and Finance},
            addressline={BITS Pilani K.K. Birla Goa Campus}, 
            state={Goa},
            postcode={403726}, 
            country={India}}

\begin{abstract} 
Medical time series datasets feature missing values that need data imputation methods, however, conventional machine learning models fall short due to a lack of uncertainty quantification in predictions.  
Among these models, the CATSI (Context-Aware Time Series Imputation) stands out for its effectiveness by incorporating a context vector into the imputation process, capturing the global dependencies of each patient. In this paper, we propose a Bayesian Context-Aware Time Series Imputation (Bayes-CATSI) framework which leverages uncertainty quantification offered by variational inference. We consider  the time series derived from electroencephalography (EEG), electrooculography (EOG), electromyography (EMG), electrocardiology (EKG).  Variational Inference assumes the shape of the posterior distribution and through minimization of the Kullback-Leibler(KL) divergence it finds variational densities that are closest to the true posterior distribution. Thus , we integrate the variational Bayesian  deep learning layers into the CATSI model. 
Our results show that Bayes-CATSI not only provides uncertainty quantification but also achieves superior imputation performance compared to the CATSI model. Specifically, an instance of  Bayes-CATSI outperforms CATSI  by 9.57\%.  We provide an open-source code implementation for applying Bayes-CATSI to other medical data imputation problems.

 
\end{abstract}


\end{frontmatter}


\section{Introduction}
\label{sec:sample1}

Electronic Health Records (EHRs) have become increasingly important in healthcare over the past decade, revolutionising the way medical information is managed and utilised \cite{Evans2016ElectronicHR}. EHRs contain a wide range of data including the medical history of patients for the analysis of medically relevant trends  ver time \cite{CHAKRABARTY201929}. This enables  the opportunity to conduct secondary analysis \cite{Cheng2014SecondaryAO} using the EHR data. Laboratory tests are essential tools for assessing and managing a patient's health status as they  provide continuous and detailed information that is crucial for effective diagnosis, treatment, and patient safety. Therefore, an important component of healthcare data analytics is modelling clinical time series \cite{Harutyunyan2017MultitaskLA, Song2017AttendAD}. Data quality is of utmost importance when using clinical time series data for decision-making; Kramer et al. \cite{KRAMER2021106359} have shown that there is a strong connection between data quality and clinical intervention that affects the physician's decision-making ability. Clinical time series data often suffers from low quality due to noise, missing values, and irregular collection intervals. These issues necessitate extensive pre-processing to ensure the data is usable for accurate clinical decision-making\cite{Xiao2018OpportunitiesAC}.  Hence, it is important to estimate missing values in a clinical time series, also known as  data imputation, where machine learning models have been successfully utilised  \cite{pmlr-v56-Lipton16}.


  Imputing missing values in clinical time series data is complex for several reasons. Firstly, clinical measurements are often recorded at irregular intervals. For example, a patient's heart rate can be monitored continuously, but blood tests are taken only when deemed necessary, leading to gaps in the data \cite{Yin2020ContextAwareTS}. Secondly, the pattern of missing data is not always random. Laboratory tests, such as blood glucose levels, might not be conducted if the caregiver assesses that the patient’s condition does not require it \cite{TAN2023104306}. This means the absence of data can indicate a patient's stable health, introducing a bias based on health state. Lastly, missing data often occurs in blocks rather than isolated points. For instance, if a patient is being transferred between hospital wards, monitoring equipment might be temporarily disconnected, resulting in a continuous segment of missing data \cite{Yin2020ContextAwareTS}. These scenarios highlight the challenges in imputing clinical time series data, as they require sophisticated methods to handle non-random and consecutive missing values effectively.

Statistical and machine learning models have been prominent for imputing missing data    \cite{Emmanuel2021,JEREZ2010105} for a wide range of problems that includes time series.  Nijman et al.  \cite{Nijman2021-vq}   reported that missing data has been poorly handled in machine learning-based predictive models. Statistical time series models such as  ARMA (autoregressive moving average)  and ARIMA  (autoregressive integrated moving average) \cite{10.1007/978-1-4684-9403-7_2} have been widely used; however, these models are linear and cannot handle the noisy and online nature of the data. Matrix completion \cite{NIPS2016_85422afb} is a statistics and machine learning strategy that involves the task of filling in the missing entries of a partially observed matrix, and has been successfully used in movie-rating and recommender systems in platforms such as Netflix  \cite{Ramlatchan2018}. Matrix completion has been used for data imputation; however,  it requires strong assumptions such as temporal regularity and low rankness. Deep learning methods such as recurrent neural networks (RNNs) have been used for data imputation \cite{NEURIPS2018_734e6bfc, Che2018} in time series. RNNs have been prominent for modelling temporal data as they use hidden states to capture past observations and provide feedback mechanisms that are not present in simple neural networks  {\cite{petneházi2019recurrentneuralnetworkstime}. A prominent  RNN implementation for time series includes the Bidirectional Recurrent Imputation for Time Series (BRITS)\cite{NEURIPS2018_734e6bfc}. A major problem faced by BRITS is that they tend to capture local properties rather than global dynamics of the input, which could have a greater impact in the case of clinical time series. For example, a patient with high blood pressure will likely have different temporal patterns compared to other patients, these patterns will be identified only by models that capture the global dynamics of the input data. The CATSI (Context-Aware Time Series Imputation) model \cite{Yin2020ContextAwareTS} was developed to address the challenge of capturing global dynamics. The CATSI model builds upon the bidirectional LSTM architecture of BRITS by incorporating a 'context vector' that captures the global dynamics of patients' input data. This context vector allows the model to account for broader temporal dependencies and correlations, leading to more accurate and contextually aware imputations. 
 
 Models such as  CATSI\cite{Yin2020ContextAwareTS} and BRITS\cite{NEURIPS2018_734e6bfc} use deep learning models, resulting in point predictions that lack uncertainty quantification. Bayesian Inference provides a mechanism for estimating unknown model parameters and quantifying uncertainty in predictions \cite{MacKay1996}. Bayesian inference represents the unknown model parameters using probability (posterior) distributions and applies methods such as variational inference\cite{doi:10.1080/01621459.2017.1285773, rezende2014stochastic} and Markov Chain Monte-Carlo  (MCMC) \cite{Borkar1953EquationOS, hastings1970monte} sampling. Bayesian inference has been widely applied in healthcare \cite{doi:10.1137/19M1248352, lin2020bayesian, jena2019bayesian} as they provide uncertainty projections in predictions using the posterior distribution as opposed to the single-point estimates in conventional machine learning and deep learning models.  Guo et al. \cite{DBLP:journals/corr/abs-1911-07572} has proposed a Bayesian Recurrent Framework model for performing imputation, leveraging the strengths of Bayesian inference to enhance the imputation process.

 Significant advancements have been made in Bayesian neural networks   and Bayesian deep learning using variational inference \cite{pmlr-v37-blundell15}, as they integrate more seamlessly with gradient-based optimization methods (such as backpropagation) compared to  MCMC sampling that faces challenges with large models. Prominentimplementation  of variational inference includes Bayes-by-backdrop \cite{pmlr-v37-blundell15} which introduces a backpropagation-compatible algorithm for learning a probability distribution on the weights of a neural network and variational autoencoders \cite{kingma2022autoencodingvariationalbayes}. Bayes-by-backdrop introduces a lower bound estimator for efficient posterior inference in the presence of continuous latent variables with intractable posterior distributions.  Kapoor et al. \cite{KAPOOR2023105654} 
     applied Bayesian variational inference-based RNNs for cyclone track and intensity prediction, demonstrating their ability to model spatiotemporal data. However, variational inference provides an approximate inference by treating the marginalization required for Bayesian inference as an optimisation problem \cite{doi:10.1080/01621459.2017.1285773, wainwright2008graphical}, rather than directly sampling from the posterior distribution as done by Markov Chain Monte Carlo  (MCMC) sampling \cite{hastings1970monte} which has challenges when it comes to large number of model parameters and hence the progress in deep learning has been slow. Despite the importance of uncertainty quantification in model predictions, there has been limited application of Bayesian deep learning in data imputation. Prior work in the domain of data imputation has been focussed on traditional deep learning models including RNN  implementations such as  CATSI and BRITS \cite{NEURIPS2018_734e6bfc, Che2018, Yin2020ContextAwareTS}, and Transformer based  models such as  SAITS  (Self-Attention Based Imputation for Time Series)\cite{Du_2023}. The advancements in variational inference and deep learning methods over the past decade inspire their application for uncertainty quantification in medicine, particularly in improving the accuracy and reliability of data imputation.

 In this paper, we enhance the performance of the CATSI model by incorporating uncertainty quantification using variational inference. We integrate customised Bayesian deep learning layers into the existing CATSI model architecture, primarily replacing all deterministic deep learning layers with Bayesian deep learning layers using  Bayes-by-backprop  \cite{pmlr-v37-blundell15}. Our novel  model, referred to as 'Bayes-CATSI,' requires computationally intensive execution, necessitating a limit on the number of samples used. To address this limitation, we   propose 'partial-Bayes-CATSI,' which replaces only a subset of deterministic deep learning layers with Bayesian deep learning layers. This variant allows us to test the impact of incorporating a limited number of Bayesian layers while using more data samples for execution. By comparing the performance of both models Bayes-CATSI and partial-Bayes-CATSI against the original CATSI model, we demonstrate the immediate improvement in results achieved by incorporating Bayesian layers. The performance of these models has been evaluated using medical data from the Computing in Cardiology Challenge 2018 \cite{Ghassemi2018YouSY}. Our results highlight the benefits of uncertainty quantification and the trade-offs between computational complexity and model performance for the imputation problem.

  The rest of the paper is organized as follows. Section 2 provides a background on related methods and Section 3 presents the data pre-processing and the proposed method. 
Section 4 presents experiments and the results, Section 5 discusses the results and Section 6 concludes with future research directions.

\section{Related Work}
\subsection{Time Series Data Imputation}

A substantial body of literature exists on time series imputation \cite{inproceedings123}, with most studies leveraging longitudinal observations \cite{kazijevs2023deepimputationmissingvalues} and cross-feature correlations \cite{LNLM2022-20013} to improve the imputation process with impositions of various assumptions. As mentioned,  autoregressive models, such as ARMA and ARIMA, are among the simplest models used for time series analysis. These models assume the input data is stationary. Although differencing \cite{RePEc:bla:jtsera:v:1:y:1980:i:1:p:15-29} can transform non-stationary data into a stationary form, this process often results in the loss of complex temporal dynamics, which can lead to inevitable errors in the model's predictions. In Bayesian data analysis, Gaussian Processes models \cite{mackay1998introduction,SCHULZ20181}  excel at handling uncertainties in observed data. However, they are very sensitive to the choice of prior, which influences the parameters of the underlying data-generating process. In clinical time series, encoding patients' rapidly changing health states as a prior distribution is particularly challenging. Additionally, Gaussian Process\cite{Hori2016MultitaskGP} typically assume that data points closer in time have more similar values due to an implicit locality constraint, which can be limiting when addressing the complex and global dependencies common in clinical time series data.
Matrix factorisation \cite{NIPS2016_85422afb} and their higher-order extension, along with tensor factorisation  \cite{CONG201559} are commonly used techniques for analysing and imputing time series data. These methods rely on the assumption that the observed data are generated by a linear combination of low-dimensional latent factors, suggesting low rankness. However, this assumption often proves inadequate for clinical time series, which typically exhibit complex and intricate temporal dynamics due to the multifaceted nature of patient health states and the influence of various physiological processes over time. Consequently, more sophisticated approaches are often required to accurately capture and model these complexities. Multiple Imputation by Chained Equations (MICE) \cite{Azur2011MultipleIB} is a widely used method for time series imputation. Unlike single imputation, which generates only one estimate for missing data, MICE creates multiple imputations to account for the inherent uncertainty in the data, resulting in more robust and reliable estimates. This approach iteratively fills in missing values by modelling each variable as a function of the others, capturing the complex relationships among variables. However, MICE operates under the assumption that data are \textit{missing completely at random} \cite{doi:10.1080/01621459.1988.10478722}. This assumption can be problematic when applied to clinical time series, where the pattern of missing data is often influenced by patients' health states, medical interventions, and other contextual factors. In clinical settings, data might be \textit{missing not at random} \cite{e4c2626352384acea0d3a60b65f754a4} and \textit{missing at random}\cite{8db34612-77bf-3ea6-8f64-d854c4cd462c} when certain tests are only conducted if specific symptoms are present \cite{kazijevs2023deepimputationmissingvalues}. 

 Recurrent Neural Networks (RNNs) have recently gained prominence for modelling sequential data \cite{hou2024rwkvtstraditionalrecurrentneural}, including their application in time series imputation \cite{Flores2020}. RNNs utilise hidden states to encapsulate past observations and predict future time steps based on current inputs, making them adept at handling temporal dependencies. RNNs are flexible models and hence the architecture and training algorithm can be altered and extended depending on the application. Che et al.\cite{Che2018} introduced specific RNN architectures tailored to exploit patterns in missing data related to underlying labels in clinical time series.  Furthermore,  Cao et al.\cite{NEURIPS2018_734e6bfc} introduced a bidirectional  LSTM model that not only considers historical data but also future trends within the time series. This approach enhances the accuracy of imputation by incorporating both past context and future trends, thereby providing a comprehensive view of temporal data dynamics. Although RNNs excel in capturing temporal relationships without imposing strict assumptions on data generation, their optimisation complexity (suitable training algorithm and model architecture) can limit their ability to learn global patterns effectively. Consequently, RNNs often excel in capturing local dependencies within sequential data but may struggle with broader, global dynamics \cite{dieng2017topicrnnrecurrentneuralnetwork}.
 
  Addressing the challenges observed in RNN-based methods, the CATSI model \cite{Yin2020ContextAwareTS} was developed to extend the architecture by  Cao et al.\cite{NEURIPS2018_734e6bfc}, leveraging bidirectional LSTM to enhance its capabilities. The CATSI model incorporates a 'context vector' to encapsulate information crucial for capturing the global dynamics inherent in clinical time series data. Unlike traditional RNNs, which focus on local dependencies, CATSI  integrates broader contextual information, thereby addressing the limitations associated with purely local modelling strategies.

\subsection{Bayesian inference with variational inference}

Bayesian inference using MCMC methods faces challenges like convergence issues and inefficiency in high-dimensional spaces, making them unsuitable for big data and large models. Strategies combining MCMC with gradient-based methods \cite{https://doi.org/10.1111/j.1467-9868.2010.00765.x, 10.1111/1467-9868.00123, neal2012mcmc, 10.5555/3104482.3104568, Chandra_2019} and meta-heuristic approaches \cite{10.1007/978-3-540-24621-3_6, 10.5555/3041838.3041931, Braak2006AMC, Braak2008DifferentialEM} have been developed. Structural changes such as nested sampling \cite{10.1214/06-BA127} and parallel tempering MCMC \cite{PhysRevLett.57.2607, doi:10.1143/JPSJ.65.1604} improve efficiency. Recent advances with parallel tempering MCMC and Langevin gradients have shown promise for Graph CNNs \cite{9535500} and deep autoencoders \cite{Chandra_2022}; however, the challenge remains as the data and model size increase.

 Variational inference offers a tractable alternative, optimising variational densities to approximate the posterior distribution and minimise Kullback-Leibler (KL) divergence. Early variational inference approaches for Bayesian neural networks (BNNs) used mean field variational Bayes (MFVB) \cite{Barber1998EnsembleLI, Hinton1993KeepingTN}.  Variational inference gained popularity with deep learning for robust uncertainty quantification. Graves et al.\cite{NIPS2011_7eb3c8be} introduced computation of derivatives of expectations in the variational
objective function also known as evidence lower bound (ELBO). Blundell et al. \cite{pmlr-v37-blundell15}  simplified the implementation of variational inference for neural networks using Bayes-by-backdrop.  As the sampling operation is not deterministic (non-differentiable), in order to perform gradient optimization over the
variational loss the re-parameterization strategy by Kingma et al.\cite{kingma2022autoencodingvariationalbayes} and Rezende et al.\cite{rezende2014stochastic} allowed the representation of random variables (i.e., trainable parameters) as deterministic functions with added noise which results in model optimisation (training) using stochastic gradient descent (SGD). Consequently, Bayes-by-backprop leverages this trick to estimate the derivatives of expectations, facilitating the training of Bayesian neural networks to incorporate weight uncertainty  crucial in balancing the trade-off between exploration and exploitation in reinforcement learning scenarios. Furthermore, Blundell et al.\cite{pmlr-v37-blundell15} provided empirical evidence showing that Bayes-by-backprop not only simplifies the training process but also delivers superior performance compared to traditional regularisation techniques like dropout. Bayes-by-backprop achieves this through a straightforward formulation of the loss function, known as the variational free energy or evidence lower bound. This formulation captures the uncertainty in model weights more effectively, leading to better generalisation and robustness in neural network predictions.
 
We did not find any study that considered merging Bayesian  variational inference into a model such as  CATSI which incorporates the global dynamics along with the local dynamics of a clinical time series using deep learning.

\section{Methodology}
\subsection{Dataset}

We utilise data from the publicly available dataset collected in real-world intensive care units (ICUs) for the Computing in Cardiology challenge 2018 \cite{Ghassemi2018YouSY}. The data were collected by Massachusetts General Hospital’s (MGH) Computational Clinical Neurophysiology Laboratory (CCNL), and the Clinical Data Animation Laboratory (CDAC). The subjects (patients) had a variety of physiological signals recorded as they slept through the night including electroencephalography (EEG), electrooculography (EOG), electromyography (EMG) and electrocardiology (EKG). Mainly we have 12 analytes which correspond to each of the physiological signals mentioned: EEG ('F3-M2', 'F4-M1', 'C3-M2', 'C4-M1', 'O1-M2', 'O2-M1'), EOG('E1-M2'), EMG('Chin1-Chin2', 'ABD', 'CHEST', 'AIRFLOW'), ECG('ECG') with frequency of 200Hz. In this study, we utilised data from 20 randomly selected patients for the training set and 10 patients for the testing set. Since the ground truth of the original missing values was unknown, additional missing data was introduced by randomly masking observations, furthermore due to computational constraints, we had to limit our study to 2000 samples for the partial-Bayes-CATSI model and 500 samples for the Bayes-CATSI model. Table \ref{tab:data_table} summarises the basic characteristics of the data containing the 12 analytes that we use in our study.

\begin{strip}

\small
 
\renewcommand{\tabcolsep}{1.1pc} 
\renewcommand{\arraystretch}{1.5} 
\begin{tabular}{|c|c|c|c|c|c|c|}
\hline
\multirow{2}{*}{\textbf{Analyte}} & \multicolumn{2}{c|}{Mean} &
\multicolumn{2}{c|}{SD} &
\multicolumn{2}{c|}{Missing Rate(\%)} \\ 

  & \textbf{Training} & \textbf{Testing} & \textbf{Training} & \textbf{Testing} & \textbf{Ground-Truth} & \textbf{After masking} \\ \hline
\textbf{F3-M2} & 0.243750 & 0.400400 & 24.933606 & 19.024624 & 0.165000 & 0.561875 \\ \hline
\textbf{F4-M1} & 0.553150 & 0.527200 & 19.600089 & 18.782545 & 0.174167 & 0.568125 \\ \hline
\textbf{C3-M2} &0.077325 & -0.037000 & 17.667076 & 14.882407 & 0.163750 & 0.566875 \\ \hline
\textbf{C4-M1} & 0.287150 & 0.130800 & 14.351527 & 13.887232 & 0.157083 & 0.542917 \\ \hline
\textbf{O1-M2} & 0.062200 & 2.665450 & 31.900061 & 19.860268 & 0.166667 & 0.581458 \\ \hline
\textbf{O2-M1} & 0.010050 & 0.266400 & 21.455265 & 12.225981 & 0.178958 & 0.586042 \\ \hline
\textbf{E1-M2} & -0.349725 & -0.086900 & 29.426300 & 19.783182 & 0.171667 & 0.569375 \\ \hline
\textbf{Chin1-Chin2} & 0.003400 & -0.000050 & 6.082936 & 10.468178 & 0.156875 & 0.545625 \\ \hline
\textbf{ABD} & 26.102925 & 19.350350 & 531.707319 & 916.411569 & 0.153750 & 0.535417 \\ \hline
\textbf{CHEST} & 25.786000 & -22.937200 & 224.839093 & 370.659810 & 0.157917 & 0.550208 \\ \hline
\textbf{AIRFLOW} & -1.445150 & 15.663700 & 250.859185 & 357.866155 & 0.168958 & 0.573125 \\ \hline
\textbf{ECG} & 0.000176 & 0.000023 & 0.166902 & 0.174157 & 0.165208 & 0.562917 \\ \hline
\end{tabular} \\[5pt]

\captionof{table}{Basic characteristics of 12 Analytes showing the Mean and the Standard Deviation (SD). We separately list the empirical mean and the SD of the Training set and the Test set.  We compute the missing rates calculated as the number of missing time steps divided by the total time steps. 
The'Ground-Truth' column refers to the missing values already present and 'After masking' refers to the artificially added individual missing values.} 

\label{tab:data_table}
\end{strip}

\subsection{Variational inference for training neural networks}

As mentioned earlier, Bayesian inference can be implemented with either MCMC sampling \cite{hastings1970monte} or variational inference \cite{pmlr-v37-blundell15}. The use of Bayesian inference to estimate (train) neural network parameters (weights and biases) enables them to be viewed as probabilistic models \cite{pmlr-v37-blundell15}. Essentially, Bayesian inference enables fixed point estimates of model parameters to be represented as probability distributions, also known as the posterior probability distribution which requires a prior probability distribution and an inference algorithm such as MCMC or variational inference.

    A Bayesian neural network model can be viewed as a probabilistic model \(P(y|x,\theta)\) where we have an input \(x \in \chi\) in  \(\chi\)  feature space and output \(y \in \gamma\) in \(\gamma\)  label space.   \(\theta\) represents the neural network model parameters. In the case of time series prediction and regression problems, we usually assume that \(\gamma\) follows a Gaussian distribution. The neural network objective function is given by \(f(x,\theta)\)  with parameters \(\theta\) for the pairing \((x,y)\) during the training process.
    
   We can learn (train) the model parameters (weights and biases)   given a set of training examples \(\mathcal{D} = (x_i, y_i)_i\) by using  the \textit{Maximum Likelihood Estimate} (MLE) of the parameters \(\hat{\theta}\) given as Equation \ref{eq:mle}.
    \begin{align}
    \hat{\theta} &= \arg \max_{\theta} \log P(\mathcal{D} \mid \theta)  \\
                 &= \arg \max_{\theta} \sum_i \log P(y_i \mid \mathbf{x}_i, \theta) \label{eq:mle}
    \end{align}

 Typically,  training of the neural network model is done using the backpropagation algorithm, which employs gradient-based optimisation. The model output \(P(y,x)\) has to be differentiable with respect to \(\theta\). We need a prior distribution \(P(\theta)\) for the parameters \(\theta\) and in the case of neural networks, Gaussian priors are typically used \cite{pmlr-v37-blundell15}. \(\hat{\theta}\)  represents the \textit{maximum a posteriori} (MAP) estimate of the parameters as shown in Equation \ref{eq:map2}.
    \begin{align}
    \hat{\theta} &= \arg \max_{\theta} \log P(\theta \mid \mathcal{D})  \label{eq:map} \\
                &= \arg \max_{\theta} \left( \log P(\mathcal{D} \mid \theta) + \log P(\theta) \right) \label{eq:map2}
    \end{align}
    The prior used has a Gaussian Distribution with zero mean and standard deviation   \(\tau\) and the prior density is  given in Equation  \ref{eq:prior}

    \begin{align}
    P(\theta) &\propto \frac{1}{(2\pi\tau^2)^{L/2}} \exp\left\{-\frac{1}{2\tau^2} \sum_{l=1}^L \theta_l \right\} \label{eq:prior}
    \end{align}

    where $L$ is the total number of trainable parameters in the neural network. We use variational inference to approximate the posterior distribution of the model parameters. We assume a variational posterior on the neural network weights given by \(q(\theta|\delta)\), parameterised by \(\delta\). Variational inference is then used to find the value of \(\delta\) that minimises the KL-divergence between the variational posterior and the true posterior as shown in Equation \ref{eq:deltaeq2}

    \begin{align}
    \hat{\delta} &= \arg \min_{\delta} \text{KL}[q(\theta \mid \delta) \parallel P(\theta \mid \mathcal{D})] \label{eq:deltaeq1} \\
                &= \arg \min_{\delta} \int q(\theta \mid \delta) \log \frac{q(\theta \mid \delta)}{P(\theta) P(\mathcal{D} \mid \theta)} \, dw \label{eq:loss_simplify}\\
                 &= \arg \min_{\delta}  \text{KL}[q(\theta \mid \delta) \parallel P(\theta)] - \mathbb{E}_{q(\theta \mid \delta)} [\log P(\mathcal{D} \mid \theta)] \label{eq:deltaeq2}
    \end{align}
    
    where \(P(\theta \mid \mathcal{D})\) is the true posterior and \(P(\mathcal{D} \mid \theta)\) is the likelihood. We present the loss function in Equation \ref{eq:lossfunc}.

    \begin{align}
    \mathcal{L} &= \text{KL}[q(\theta \mid \delta) \parallel P(\theta)] - \mathbb{E}_{q(\theta \mid \delta)} \left[ \log P(\mathcal{D} \mid \theta) \right] \label{eq:lossfunc}
    \end{align}

    The loss function is thus the sum of a data dependent part referred to as the likelihood cost and a prior-dependent part referred to as the complexity cost which requires calculation of the KL-Divergence. The loss function thus has to undergo a trade-off between satisfying the complexity of the data and satisfying the simplicity of the prior \(P(\theta)\). We   use the formulation in Blundell et al.\cite{pmlr-v37-blundell15} to approximate the loss function in Equation \eqref{eq:lossfunc} using sampling given by:
    
    \begin{align}
    \mathcal{L} &\approx \sum_{i=1}^m \left( \log q(\theta^{(i)} \mid \delta) - \log P(\theta^{(i)}) P(\mathcal{D} \mid \theta^{(i)}) \right) \label{eq:simpleLoss} \\
    &\approx \sum_{i=1}^m \left( \log q(\theta^{(i)} \mid \delta) - \log P(\theta^{(i)})\right) 
    - \sum_{i=1}^m \left( \log P(\mathcal{D} \mid \theta^{(i)}) \right)
    \label{eq:simpleLoss2}
    \end{align}

    where \(\theta^{(i)}\) represents the \(i\)th sample drawn from the variational posterior \(q(\theta \mid \delta)\). The former part of the Equation \eqref{eq:simpleLoss2} refers to the complexity cost, while the latter refers to the likelihood cost described earlier.

\subsection{Framework}

 This section presents the   Bayes-CATSI and partial Bayes-CATSI framework that features variational inference and deep learning models.


\subsubsection{Pre-processing: masking, normalisation and temporal decay}

We follow data input and masking procedures by Yin et al.\cite{Yin2020ContextAwareTS} and Cao et al.\cite{NEURIPS2018_734e6bfc} where artificial missing values were introduced into the input data, which already contains pre-existing missing values. The goal of introducing the artificial missing values is to generate a test dataset for evaluating the model. Figure \ref{fig:general_process} provides an overview of the framework for the imputation process, where we use the incomplete data, along with the observation mask and evaluation mask for the imputation model. 

We represent the input data corresponding to the multivariate time series for a patient as \( \mathbf{X} \in \mathbb{R}^{T \times F} \), where \( T \) is the number of time steps and \( F \) is the number of features (or variables) measured at each time step and the \( t \)-th row \( \mathbf{x}_t \) is the observation at the \( t \)-th time step. \( \mathbf{s}_t \)  denotes the timestamp corresponding to the \( t \)-th time step. As demonstrated in the existing work \cite{Yin2020ContextAwareTS, NEURIPS2018_734e6bfc}, we use a masking matrix or the observation matrix \textbf{M} with the same size of multivariate time series \( \mathbf{X} \) to indicate the missing data in the time series as shown in Equation \ref{eq:missingmask} \textcolor{black}{for example, the masks would have 500 rows and 12 columns for Bayes-CATSI and 2000 rows and 12 columns for partial Bayes-CATSI}.

\begin{equation}
     m_{t}^{f} = \begin{cases} 
    1 & \text{if the } f\text{-th variable is observed at time  \( \mathbf{s}_t \) } \\
    0 & \text{otherwise}
    \end{cases}
\label{eq:missingmask}
\end{equation}

We need to account for the irregularity of the time series caused by the missing values. Yin et al.\cite{Yin2020ContextAwareTS} introduced an observation gap matrix \( \mathbf{\Delta} \) with the same size as that of \( \mathbf{X} \) to represent the current time stamp and the time stamp of the last observation that is not missing.  We show the calculation of the  observation matrix \(\mathbf{\Delta}\)  in Equation \eqref{eq:delta_function}.

\begin{equation}
\delta_{t}^{f} = \begin{cases}
    s_t - s_{t-1} + \delta_{t-1}^{f} & \text{if } t > 1, \, m_{t-1}^{f} = 0 \\
    s_t - s_{t-1} & \text{if } t > 1, \, m_{t-1}^{f} = 1 \\
    0 & \text{if } t = 1
\end{cases}
\label{eq:delta_function}
\end{equation}

The raw input data for each variable of each patient is first normalised using min-max normalisation, as defined by Equation \ref{eq:normalized_variable}.

\begin{equation}
x^f = \frac{x^f - \min(x^f)}{\max(x^f) - \min(x^f)}
\label{eq:normalized_variable}
\end{equation}

We transform the data into a pre-completed version \(\Tilde{X}\) to effectively utilise the raw input data containing missing values in the model. \textcolor{black}{As described by Yin et al. \cite{Yin2020ContextAwareTS} and originally introduced by Che et al. \cite{Che2018}, this transformation is achieved by imputing the missing entries using a trainable temporal decay factor as depicted in Equation s\eqref{eq:gamma_t} and \eqref{eq:x_dt}. As explained by Yin et al. \cite{Yin2020ContextAwareTS}, the strategy stems from an observation that if a variable is unobserved for a long time, it gravitates towards a "default" value (the empirical mean in this case); otherwise, it stays near its historical observations.}
We use the observation \(\delta_{t}\) gap at time \(s_t\)  to compute the temporal decay factor \(\gamma_t \in \mathbb{R}^{F} \) to compute the pre-completion in Equation \ref{eq:gamma_t}.

\begin{equation}
\gamma_t = \exp \left\{ -\max(0, W_\gamma \delta_t + b_\gamma) \right\}
\label{eq:gamma_t}
\end{equation}

\begin{equation}
x'^{f}_{t} = \gamma_{t}^{f} x^{f}_{t'} + (1 - \gamma_{t}^{f}) \bar{x}^f
\label{eq:x_dt}
\end{equation}

where \(x'^{f}_{t}\) is the computed pre-completion, \(x^{f}_{t'}\) is the last non-missing observation and \(\bar{x}^f\) is the empirical mean for the \( f \)-th variable at time \(s_t\). 
We use Equation (\ref{eq:x_dt}) to replace the missing values in the raw input while preserving the observed ones.  Followed by this, we generate the pre-completed input as shown in Equation \eqref{eq:x_tilda}, as depicted in Figure \ref{fig:detailed_overview} we feed this pre-completed input into Bayes-CATSI and Partial Bayes-CATSI.

\begin{equation}
\Tilde{x}^{f}_{t} = m_{t}^{f} x^{f}_{t} + (1 - m_{t}^{f}) x'^{f}_{t}
\label{eq:x_tilda}
\end{equation}

We update the temporal decay module's parameters \(W_\gamma \in \mathbb{R}^{F \times F}\) and  \(b_\gamma \in \mathbb{R}^{F}\) during training. Figure \ref{fig:detailed_overview} demonstrates the pre-processing workflow, where the temporal decay module converts the raw input into pre-completed input \textcolor{black}{by taking the convex combination of the
 last observation and the mean with the decay factor as the coefficient, this is done} for Bayes-CATSI and Bayes-CATSI$_{partial}$  models.

\begin{figure*}[htbp!]
  \centering
  \includegraphics[width=0.8\linewidth]{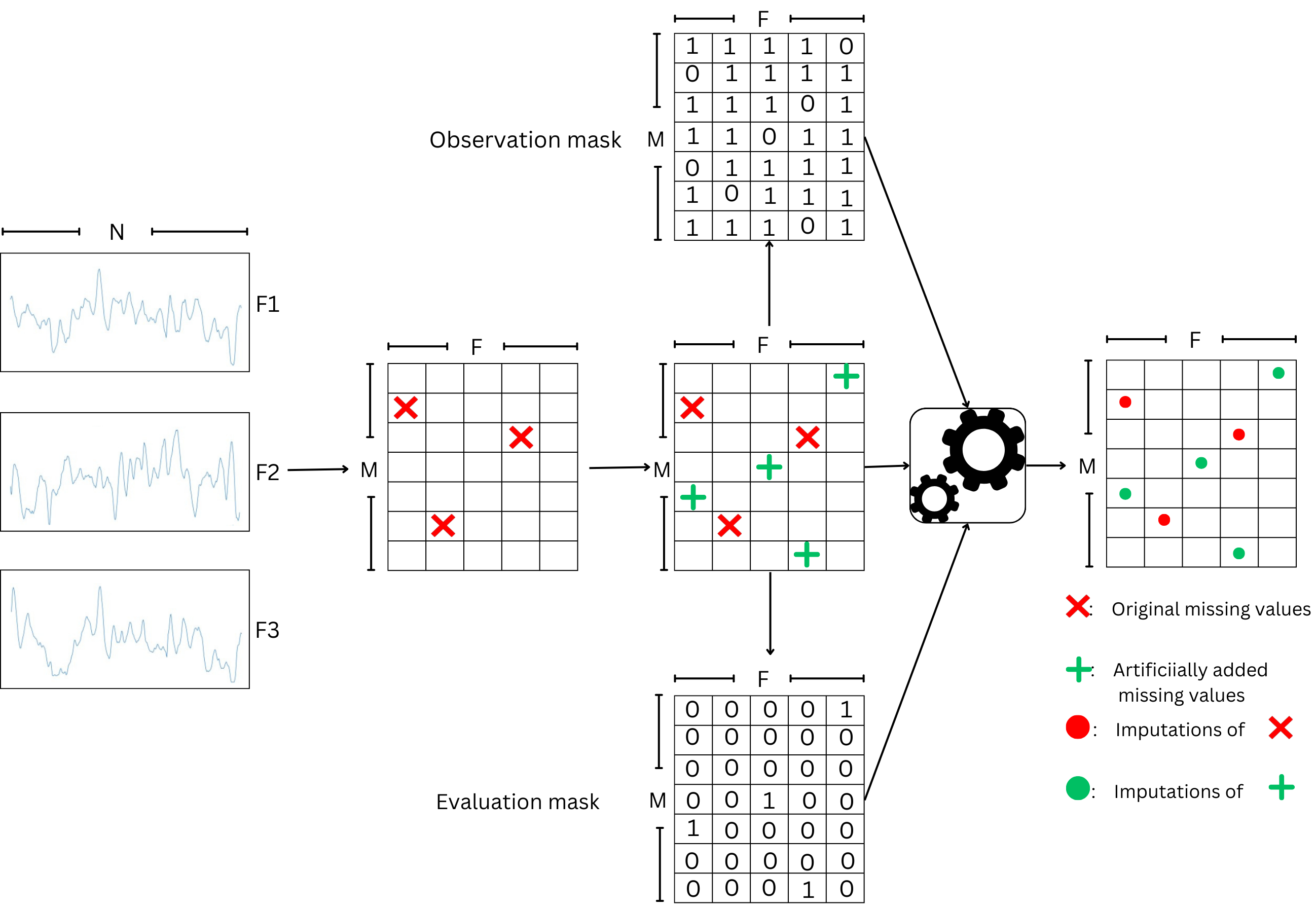}
  \captionof{figure}{A graphical overview of the imputation process  showing the masking procedure from the raw input data through which observation mask and evaluation mask are generated and fed into the imputation model. We show the data loaded into the matrices  in the left side,  wherein $N$ time steps on the x-axis correspond to the $M$ rows in the matrix, while F1, F2, F3 correspond to the first 3 features/columns in the matrix. We mark the missing values in the input data  as crosses in the matrix corresponding to the particular feature and time step and the 'gear box' depicts the imputation model. We present  the final output to give a general idea of the input and output of the imputation model. All matrices depicted have dimensions $M$ by $F$, where $F$ represents  features of the multivariate time series. The red 'X' indicates the original missing values in the input dataset. The green '+' indicates the missing value deliberately added to evaluate the imputations conducted by the model. The observation mask stores the locations of the values that are present by labelling them as 1 and values that are absent by labelling them as 0. Similarly, the evaluation mask stores locations of missing values that have been deliberately added by labelling them as 1 and labelling other values as 0. These masks help the model differentiate between original missing values, deliberately added missing values and non-missing values.}
  \label{fig:general_process}
\end{figure*}

\subsubsection{Bayes-CATSI and Partial Bayes-CATSI}

The imputation model by Yin et al. \cite{Yin2020ContextAwareTS}  shown in Figure \ref{fig:detailed_overview} features three key components including context-aware recurrent imputation, cross-feature imputation, and a fusion layer that integrates the results from the preceding layers to generate the final imputed dataset. Our Bayes-CATSI and partial Bayes-CATSI models include modifications to the internal layers of these key components, with further details available in the subsequent sections.

\subsubsection{Context-Aware Recurrent Imputation}

 Figures \ref{fig:detailedBayesCATSI} and \ref{fig:detailedPartialBayesCATSI} provide a detailed diagram of the architectures of Bayes-CATSI and Partial Bayes-CATSI where the "Context Aware Recurrent Imputation" component includes a "context vector" represented as \(r \in \mathbb{R}^{C}\) in its imputation process. We optimise (train) this context vector \(r\)  to extract the global temporal dynamics to represent the corresponding \textcolor{black}{sample data's global characteristics for e.g. a patient's health state} where $C$ is the dimension of the context vector which is a pre-specified hyperparameter. 

Yin et al. \cite{Yin2020ContextAwareTS} utilised a bidirectional RNN model for imputation by feeding the time series into the forward RNN in original order and into the backward RNN in reverse. The essence of using a bidirectional RNN is that at time step \(t\) the hidden states before and after \(s_t\) are computed using forward and backward RNN. This is followed by the recurrent imputation which combines the two hidden states by applying a linear transformation as shown in Equation \eqref{eq:x_hat_t}
\begin{equation}
\hat{x}_t = W_x [
\overrightarrow{h}_{t-1};
\overleftarrow{h}_{T-t} 
] + b_x
\label{eq:x_hat_t}
\end{equation}
where \(h\) is the hidden state, and [;] indicates concatenation of two hidden states. \(\overrightarrow{h}\) with a rightward arrow atop denotes the parameters of the forward RNN while \(\overleftarrow{h}\) with a leftward arrow atop denotes the parameters of the backward RNN. 

The modification that we make to the original CATSI model is that both the Bayes-CATSI and partial Bayes-CATSI models (Figures \ref{fig:detailedBayesCATSI} and \ref{fig:detailedPartialBayesCATSI}) utilise a Bayesian linear layer to combine the forward and backward hidden states, making the weights \(W_x\) and biases \(b_x\) in Equation \ref{eq:x_hat_t} probabilistic, rather than deterministic (as in original CATSI).
\begin{figure*}[htbp!]
  \centering
  \includegraphics[width=0.8\linewidth]{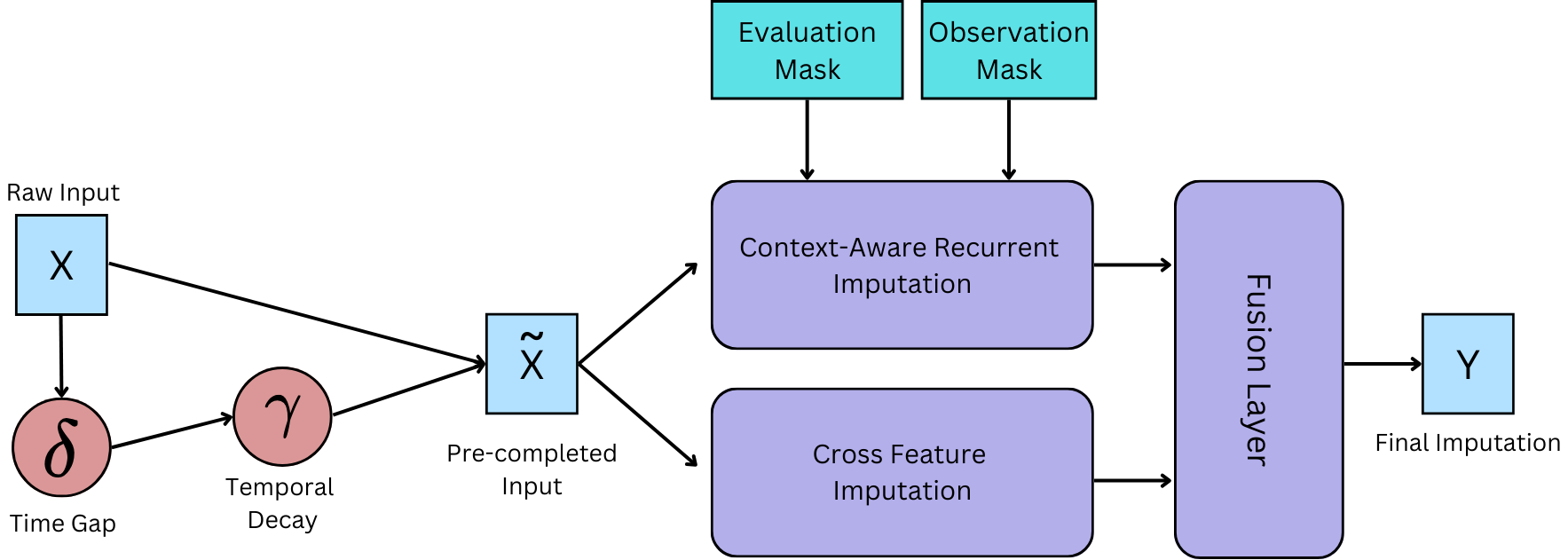}
  \captionof{figure}{Overview of the internal processing of the imputation model showing the generation of the pre-completed input from the raw input.}
  \label{fig:detailed_overview}
\end{figure*}
\begin{figure*}[htbp!]
  \centering
  \includegraphics[width=0.9\linewidth, height = 0.4\textheight]{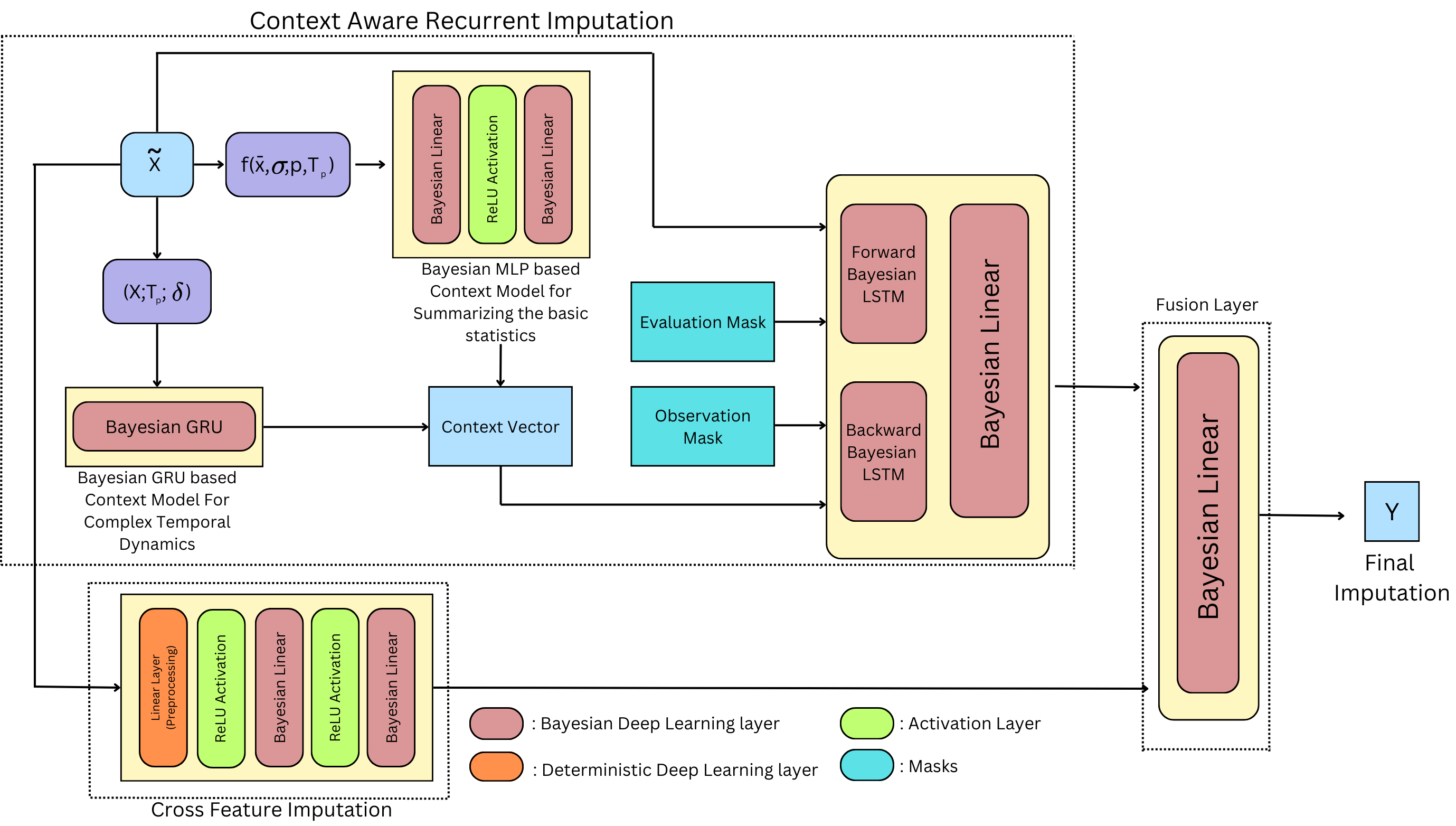}
  \captionof{figure}{Detailed architecture of the Bayes-CATSI model}
  \label{fig:detailedBayesCATSI}
\end{figure*}

\begin{figure*}[htbp!]
  \centering
  \includegraphics[width=0.9\linewidth, height = 0.4\textheight]{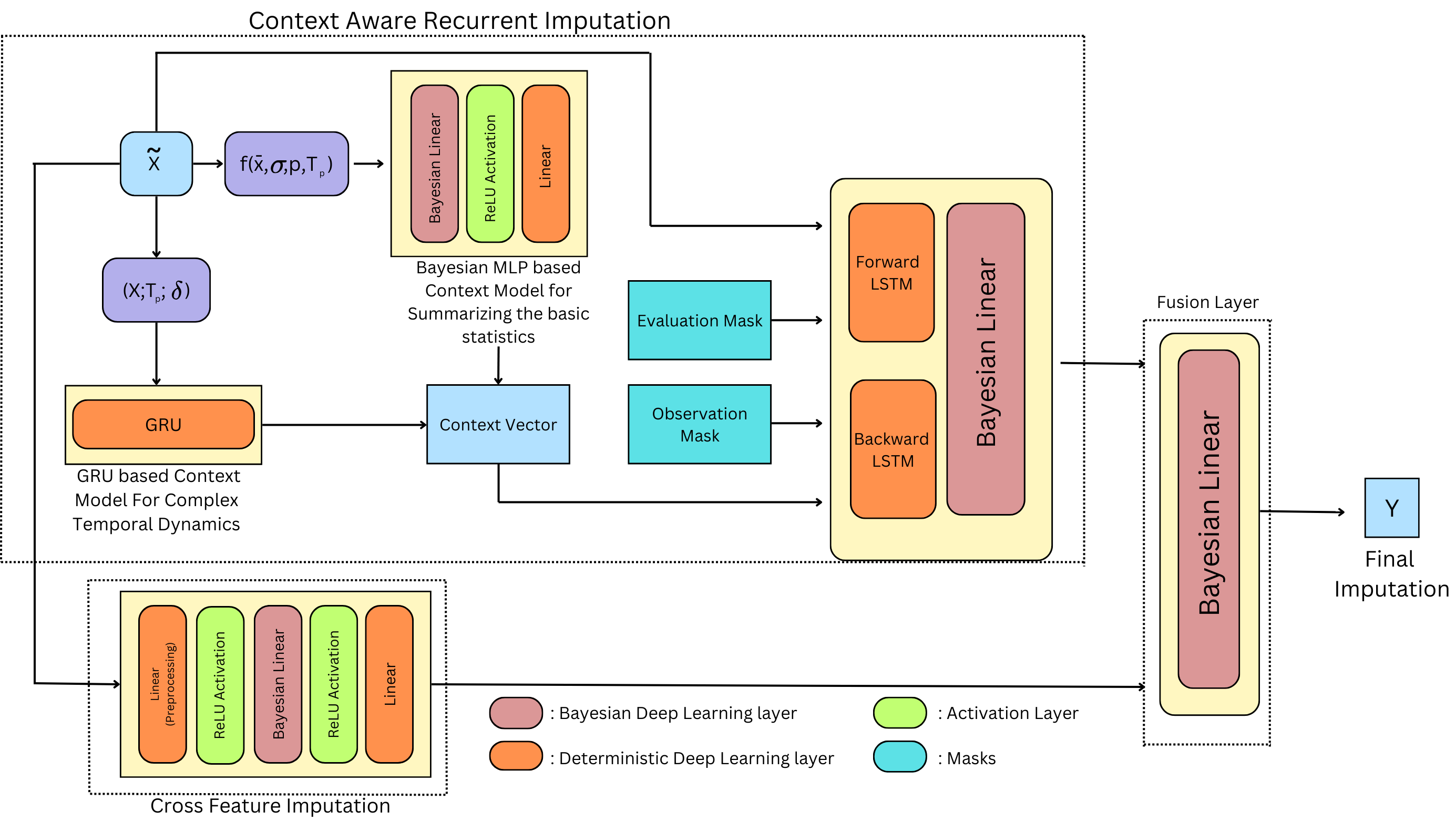}
  \captionof{figure}{Detailed architecture of the Partial Bayes-CATSI model}
  \label{fig:detailedPartialBayesCATSI}
\end{figure*}

Yin et al. \cite{Yin2020ContextAwareTS} in CATSI  employed the LSTM model to extract hidden states \(h\) from the input data and the context vector. The input data alone would neglect the patient's health states, hence  CATSI used a context vector as described earlier which helps incorporate patients' health states or global temporal dynamics of the corresponding patient. At each time step, the hidden states are produced with reference to the global dynamics of the input time series by feeding the context vector to the LSTM model together with the time series input. In Bayes-CATSI, we use a probabilistic (Bayesian) LSTM layer while in  partial Bayes-CATSI, we use a conventional LSTM model with deterministic weights due to the computationally intensive nature of the Bayesian LSTM layer. 

\begin{equation}
\overrightarrow{h}_{0} = 
\overrightarrow{W}_{h} r + \overrightarrow{b}_{h},  
\overrightarrow{c}_{0} = tanh(\overrightarrow{h}_{0})
\label{eq:h0_forward}
\end{equation}

\begin{equation}
\overleftarrow{h}_{0} = \overleftarrow{W}_{h} r + \overleftarrow{b}_{h},  \overleftarrow{c}_{0} = tanh(\overleftarrow{h}_{0})
\label{eq:h0_backward}
\end{equation}

The weights \(\overrightarrow{W}_{h}\) and biases \(\overrightarrow{b}_{h}\) in Equation \ref{eq:h0_forward} and \(\overleftarrow{W}_{h}\) and biases \(\overleftarrow{b}_{h}\) in Equation \ref{eq:h0_backward} are probabilistic in case of the Bayes-CATSI model, while they are deterministic in case of the partial Bayes-CATSI model. The context vector \(r\) is used to initialise the hidden state \(h_0\) and cell state \(c_0\) of the subsequent LSTM model, this is beneficial for imputing missing values at the first several time steps according to the original CATSI model \cite{Yin2020ContextAwareTS}. We use the Bayesian LSTM model as mentioned earlier in Bayes-CATSI while the LSTM model is used in partial Bayes-CATSI.
\begin{align}
\overrightarrow{h}_{t-1}, \overrightarrow{c}_{t-1} &= \text{LSTM}([\tilde{x}_{t-1}; r], \overrightarrow{h}_{t-2}, \overrightarrow{c}_{t-2}) 
\label{eq:partial_forward_lstm} \\
\overleftarrow{h}_{T-t}, \overleftarrow{c}_{T-t} &= \text{LSTM}([\tilde{x}_{t+1}; r], \overleftarrow{h}_{T-t-1}, \overleftarrow{c}_{T-t-1}) 
\label{eq:partial_backward_lstm}
\end{align}

Equations \ref{eq:partial_forward_lstm} and \ref{eq:partial_backward_lstm} correspond to the partial Bayes-CATSI model and Equations \ref{eq:complete_forward_lstm} \ref{eq:complete_backward_lstm} are used for Bayes-CATSI.

\begin{align}
\overrightarrow{h}_{t-1}, \overrightarrow{c}_{t-1} &= \text{Bayesian-LSTM}([\tilde{x}_{t-1}; r], \overrightarrow{h}_{t-2}, \overrightarrow{c}_{t-2}) \label{eq:complete_forward_lstm} \\
\overleftarrow{h}_{T-t}, \overleftarrow{c}_{T-t} &= \text{Bayesian-LSTM}([\tilde{x}_{t+1}; r], \overleftarrow{h}_{T-t-1}, \overleftarrow{c}_{T-t-1}) \label{eq:complete_backward_lstm}
\end{align}

 Yin et al.\cite{Yin2020ContextAwareTS} in CATSI proposed two approaches to train the context vector that included basic statistics and RNN-based model.\textcolor{black}{The first approach captures the basic statistics using an MLP layer while the second approach uses an RNN based model to capture the complex temporal dynamics.} 

 In summarising the basic statistics, CATSI captures the overall health state of patients by calculation of the basic descriptive statistics like the mean \(\bar{x}\), standard deviation \(\sigma\), missing rate \(p\) and length of the time series \(T_p\) for the given patient.

\begin{equation}
r_{MLP} = f(\bar{x}, \sigma\, p, T_p)
\label{eq:basicStats}
\end{equation}

A multilayer perception (MLP) is then used to approximate the function \(f\) to summarise the patient's overall health state. As shown in Figure \ref{fig:detailedBayesCATSI}, we use two Bayesian linear layers separated by a ReLU activation layer in the MLP-based context model in Bayes-CATSI for summarising the basic statistics. As described in Figure \ref{fig:detailedPartialBayesCATSI} we use only one  Bayesian linear layer in the partial Bayes-CATSI model to reduce the computational complexity. The MLP model however summarises only the basic characteristics of a patient's health data and may not be able to capture the complex temporal dynamics.


We use the RNN-based modelto capture complex temporal dynamics in CATSI where  the hidden states encode the temporal dynamics of the data}. We use the Gated Recurrent Unit (GRU) \cite{chung2014empiricalevaluationgatedrecurrent} model, which is a simple implementation of the LSTM  in partial Bayes-CATSI and  Bayes-CATSI.

The output from the MLP-based context model and the GRU-based context model is concatenated to output the context vector \(r\) as shown in Equation \ref{eq:context_vector}.

\begin{equation}
r = [r_{MLP};r_{GRU}] 
\label{eq:context_vector}
\end{equation}

\subsubsection{Cross-Feature Imputation}

 The context-aware recurrent imputation described in the previous section focuses on the temporal dynamics of the corresponding analytes/features.  It does not focus on the dynamic correlations that exist between different features/analytes. The cross-feature imputation enables  feature correlation, thus the value of one variable can be estimated based on other variables, simultaneously.  Equation \ref{eq:vf_t} computes a linear transformation of the raw input with the diagonal parameters of \(W_v^f\) set as zeroes to avoid \(\tilde{x}_t^f\) from predicting itself.

\begin{align}
v_t^f &= W_v^f \tilde{x}_t + b_v^f \label{eq:vf_t} \\
\hat{z}_t^f &= BayesianMLP(v_t^f) \label{eq:z_hat_t}
\end{align}

where \(\hat{z}_t^f\) is the cross-feature imputation of the \(f\)-th feature at \(t\)-th time step. The linear (pre-processing) layer in Figures \ref{fig:detailedBayesCATSI} and \ref{fig:detailedPartialBayesCATSI} implements  Equation \ref{eq:vf_t}. Following this, a probabilistic (Bayesian) MLP is used to generate the cross-feature imputations, that explore the complex feature correlations. One Bayesian layer is used for the partial Bayes-CATSI model while both linear layers are replaced by their Bayesian counterparts in the Bayes-CATSI model.

\subsubsection{Fusion Layer}

After obtaining the results from the recurrent and cross-feature imputation, both results are fused using the \(\beta_{t} \in \mathbb{R}^{F}\) which is calculated using the missing patterns in the data using missing masks \(m_t\) and observation time gaps \(\gamma_t\) as shown in Equation \ref{eq:beta_t}.

\begin{equation}
\beta_t = \text{sigmoid}(W_\beta [\gamma_t; m_t] + b_\beta) 
\label{eq:beta_t}
\end{equation}

Equation \ref{eq:yt} presents the calculation of the  final output 

\begin{equation}
y_t = \beta_t \odot \hat{z}_t + (1 - \beta_t) \odot \hat{x}_t
\label{eq:yt}
\end{equation}

where \(\odot\) indicates an element-wise matrix multiplication, \(\hat{z}_t\) indicates the cross-feature imputation and \(\hat{x}_t\) indicates the recurrent imputation and \(y_t\) is the final imputation.

\subsubsection{Bayes-CATSI: variational inference for imputation}

\begin{algorithm*}
\caption{Bayes-CATSI: variational inference for  time series data imputation}
\label{alg:process}
\begin{algorithmic}
\State \textbf{Data:} Electronic health record time series  dataset
\State \textbf{Result:} Posterior distribution of  neural network parameters $P(\theta \mid \mathcal{D})$
\State \textbf{Define} Bayesian neural network model, $y = f(\mathbf{x}, \theta)$
\State \textbf{Initialise} the variational parameters, $\delta = (\mu, \rho)$
\State \textbf{Set} Hyperparameters: $N_{\text{epoch}}$, $N_{\text{sample}}$, and $\tau$
\State \textbf{Preprocess} data into input-output matrices (Figure 1) after appropriate masking procedure required for data imputation: $\mathcal{D}: (\mathbf{X}, \mathbf{Y})$
\For{$n = 1$ to $N_{\text{epoch}}$}
    \For{$i = 1$ to $N_{\text{sample}}$}
        \State \textbf{Sample noise:} $\epsilon \sim \mathcal{N}(0, I)$
        \State \textbf{Compute the parameters:} $\theta^{(i)} = \mu + \log(1 + \exp(\rho)) \cdot \epsilon$
    \EndFor
    \State \textbf{Compute variational loss:}
    \begin{align}
    \mathcal{L} = \sum_{i=1}^{N_{\text{sample}}} \left( \log q(\theta^{(i)} \mid \delta) - \log P(\theta^{(i)}) P(\mathcal{D} \mid \theta^{(i)}) \right)
    \end{align}
    \State \textbf{Compute gradients:} $\Delta \mu$ and $\Delta \rho$
    \State \textbf{Update the variational parameters:}
    \begin{align}
    \mu &= \mu - \gamma \Delta \mu \\
    \rho &= \rho - \gamma \Delta \rho
    \end{align}
\EndFor
\end{algorithmic}
\end{algorithm*}
We next present the details of implementing variational inferences for training Bayes-CATSI. 
Algorithm \ref{alg:process} presents the training of the Bayes-CATSI and partial Bayes-CATSI models using variational inference. First, we define the Bayesian neural network  $y = f(\mathbf{x}, \theta)$, as shown in Figure \ref{fig:detailedBayesCATSI} for Bayes-CATSI and Figure \ref{fig:detailedPartialBayesCATSI} for partial Bayes-CATSI. We then initialise the variational inference parameters $\delta = (\mu, \rho)$ and set the hyperparameters, which include the number of epochs $N_{\text{epoch}}$, the number of samples for Monte Carlo sampling $N_{\text{sample}}$ (similar to training epochs in backpropagation algorithm), and the standard deviation of the scaled mixed prior $\tau = (\sigma_1, \sigma_2)$. In this case,  $\sigma_1$ and $\sigma_2$ correspond to the standard deviations of the two Gaussian densities considered for the scaled mixed prior, as described by Blundell et al.\cite{pmlr-v37-blundell15}. 

Next, we perform masking on the output matrices (Figure 1), retaining only the values that need to be imputed \cite{Yin2020ContextAwareTS}. This is done using the observation mask, which contains details of all the values that are originally missing (described as 'ground-truth' in Table \ref{tab:data_table}), along with the artificially added missing values. Additionally, the evaluation mask contains details of only the artificially added missing values and is used to evaluate the trained model. 

This masking procedure is followed by the preprocessing of the data into input-output matrices (Figure 1), which is then followed by training the model. During training, we add sample noise from a standard normal distribution and compute the model parameters corresponding to each particular sample in the epoch. The standard deviation parameter $\sigma = \log(1 + \exp(\rho))$ uses a softplus function $f(x) = \log(1 + \exp(\rho))$, which ensures that the value of $\sigma$ is always positive. After extracting the model parameters for all $N_{\text{sample}}$, we calculate the variational loss, which is further used in the backpropagation process to update the $\mu$ and $\rho$ values.

\subsubsection{Loss function}
We compute the simplified form of the loss function as described in Equation \eqref{eq:simpleLoss2}   using the likelihood cost and the complexity cost. The complexity cost is calculated using the log of prior and the variational density. The variational density is calculated by summing log probability over the Gaussian distributions parameterised by the mean and variance values given in the weight matrix. While the previous sections provide details of the loss function as a whole, this section elaborates on the likelihood cost used in the paper.  We use the mean squared deviation (MSD) of the observed entries as the likelihood cost. Yin et al.\cite{Yin2020ContextAwareTS} defines this function as:
    \begin{equation}
    L(Y) = \frac{\| M \odot (X - Y) \|_F^2}{\| M \|_F^2}
    \label{eqn:loss_function}
    \end{equation}
 where $X$ is the input data, $Y$ is the produced imputation, $M$ is the observation mask described earlier where observed values correspond to 1 and unobserved to 0; thus, square of the frobenius norm of $M$ corresponds to the number of observed data points in the input data $X$.

 As done in Yin et al.\cite{Yin2020ContextAwareTS}, we accumulate the loss for the recurrent imputations $\hat{X}$, cross-feature imputation $Z$ and the final imputation $Y$ to derive the overall likelihood cost function as shown in Equation \ref{eq:loss}:
    \begin{align}
    l &= L(Y) + L(\hat{X}) + L(Z) \\
    &= \frac{\| M \odot (Y - X) \|_F^2}{\| M \|_F^2} + \frac{\| M \odot (\hat{X} - X) \|_F^2}{\| M \|_F^2} + \frac{\| M \odot (Z - X) \|_F^2}{\| M \|_F^2}
    \label{eq:loss}
    \end{align}
 We use the Adam optimizer \cite{kingma2017adammethodstochasticoptimization} for training the respective models. 

\subsection{Accuracy metrics }

Yin et al.\cite{Yin2020ContextAwareTS} measure the performance of their model using a measure they call normalised (n) root-mean-squared error ($n$RMSE) defined in Equation \ref{eqn:rmsd}

\begin{equation}
\text{$n$RMSE}(f) = \sqrt{
\frac{
\sum_{p,t} (1 - m_{p,t}^f) \left( \frac{\left| x_{p,t}^f - y_{p,t}^f \right|}{\max\left( y_{p}^f \right) - \min\left( y_{p}^f \right)} \right)^2
}{
\sum_{p,t} (1 - m_{p,t}^f)
}}
\label{eqn:rmsd}
\end{equation}

where $p$, $f$, and $t$ indicate the indices of the patient, the analyte and the time step, respectively, and $x$ and $y$ indicate the ground truth and imputations respectively. We use this $n$RMSE in our paper to evaluate the model performance.

\section{Results}

\subsection{Experiment setup}

We begin by imputing individual missing values, initially adding artificial missing values of length one. Subsequently, we increase the length of consecutive missing values, adding artificial missing values with lengths ranging from two to five.  Hence, we generate 5 datasets with one corresponding to individual missing values and other 4 corresponding to consecutive missing values of lengths ranging from two to five. Table \ref{tab:data_table} provides further details of the missing rates for individual missing value datasetst corresponding to 2000 samples. Table \ref{tab:consecutivepartialDataDesc} and \ref{tab:consecutiveFullDataDesc} provide details of missing rates for consecutive missing values corresponding to 2000 samples and 500 samples respectively. We train the Bayes-CATSI model on 80\% of the patients and use the remaining 20\% from the training set as a validation set, which contributes to the validation loss per epoch.  

As previously discussed, Bayesian models operate on the principle of sampling, where multiple samples are drawn from the distributions of the weights, and the loss function is calculated. This loss function is then used to train the parameters of the weight distributions. In contrast, the original CATSI model, which does not focus on uncertainty quantification, does not employ this sampling method. In order to provide comparison, we use the conventional frequentest style with 30 experimental runs with different initial parameters (weights and biases) for CATSI. We then compare CATSI with Bayes-CATSI where we generate \textcolor{black}{30} predictions from posterior distribution (of weights and biases) after the sampling process (training process). This allows us to calculate the mean and standard deviation of the RMSE values in both cases. Hence, we report the results  that provide the mean ($\mu$)  and the standard deviation ($\sigma$) of the RMSE values.

\subsection{Missing data imputation --individual missing values}


We added 5\% artificially missing values to the data, which already contains pre-existing missing values, using random masking as shown in Figure \ref{fig:general_process}. 
The model's performance is then evaluated on the testing set. The results of the Bayes-CATSI model and the partial Bayes-CATSI model are compared with the original CATSI model proposed by Yin et al.\cite{Yin2020ContextAwareTS}. 

We first compare the partial Bayes-CATSI model with the CATSI model using a dataset containing 2000 samples. Afterwards, we compare the Bayes-CATSI model with the CATSI model using a dataset containing 500 samples. We were restricted to 500 samples due to limited computational power, as Bayes-CATSI requires higher computational resources due to higher number of Bayesian hidden layers requiring larger number of associated parameters (weights and biases) to be sampled by variational inference algorithm. Consequently, we present the results for the partial Bayes-CATSI model on the complete dataset, and then present the results for the Bayes-CATSI model  on a subset of the dataset used in the former case.

\begin{table*}[htbp!]
\centering
\small
\begin{tabular}{c|ccc}
\hline
Analyte & Partial Bayes-CATSI & CATSI & Mean RMSE Relative Improvement \\
\hline
\textbf{F3-M2}  & $0.1502 ~(0.0008)$ & $0.1349 ~(0.0136)$ & -10.19\% \\
\textbf{F4-M1}  & $0.1379 ~(0.0005)$ & $0.1534 ~(0.0167)$ & 10.10\% \\
\textbf{C3-M2}  & $0.1240 ~(0.0006)$ & $0.1152 ~(0.0130)$ & -7.10\% \\
\textbf{C4-M1}  & $0.1524 ~(0.0008)$ & $0.1289 ~(0.0158)$ & -15.42\% \\
\textbf{O1-M2}  & $0.1565 ~(0.0004)$ & $0.1513 ~(0.0133)$ & -3.32\%\\
\textbf{O2-M1}  & $0.1334 ~(0.0005)$ & $0.1336 ~(0.0128)$ & 0.15\% \\
\textbf{E1-M2}  & $0.1677 ~(0.0006)$ & $0.1641 ~(0.0109)$ & -2.15\%\\
\textbf{Chin1-Chin2}  & $0.1194 ~(0.0005)$ & $0.1152 ~(0.0102)$ & -3.52\%\\
\textbf{ABD}  & $0.2633 ~(0.0005)$ & $0.2519 ~(0.0100)$ & -4.33\% \\
\textbf{CHEST}  & $0.2441 ~(0.0003)$ & $0.2593 ~(0.0168)$ & 5.86\%\\
\textbf{AIRFLOW} & $0.2497 ~(0.0005)$ & $0.2469 ~(0.0155)$ & -1.12\%\\
\textbf{ECG} & $0.2087 ~(0.0012)$ & $0.1735 ~(0.0173)$ & -16.87\%\\
\textbf{Mean of mean RMSE} & 0.1756 & 0.1690 & -3.76\%\\

\hline

\end{tabular}
\caption{Partial Bayes-CATSI: Results for the test dataset for individual missing values showing RMSE mean and 95\% confidence interval for different data streams (analyte). We report 30 independent experimental runs (CATSI) and  30  posterior samples (Partial Bayes-CATSI). Mean RMSE Relative Improvement gives an idea of the performance improvement between CATSI and Partial Bayes-CATSI per analyte where the negative sign shows that Partial Bayes-CATSI performs worse than CATSI, while positive sign shows that Partial Bayes-CATSI performs better than CATSI.}
\label{tab:indMissing}
\end{table*}

\begin{table*}[htbp!]
\centering
\small
\begin{tabular}{c|ccc}
\hline
Analyte & Bayes-CATSI & CATSI & Mean RMSE Relative Improvement \\
\hline
\textbf{F3-M2}  & $0.1782~(0.0006)$ & $0.1940~(0.0237)$ & 8.14\% \\
\textbf{F4-M1}  & $0.2081~(0.0006)$ & $0.2262~(0.0212)$ & 8.00\% \\
\textbf{C3-M2}  & $0.2073~(0.0006)$ & $0.1622~(0.0241)$ & -21.76\% \\
\textbf{C4-M1}  & $0.1807~(0.0007)$ & $0.1701~(0.0277)$ & -5.87\% \\
\textbf{O1-M2}  & $0.2385~(0.0008)$ & $0.2269~(0.0222)$ & -4.86\% \\
\textbf{O2-M1}  & $0.1609~(0.0004)$ & $0.1873~(0.0279)$ & 14.10\% \\
\textbf{E1-M2}  & $0.2202~(0.0007)$ & $0.2379~(0.0179)$ & 7.44\% \\
\textbf{Chin1-Chin2}  & $0.1409~(0.0008)$ & $0.1501~(0.0255)$ & 6.13\% \\
\textbf{ABD}  & $0.2674~(0.0004)$ & $0.2995~(0.0190)$ & 10.72\% \\
\textbf{CHEST}  & $0.2467~(0.0005)$ & $0.2773~(0.0252)$ & 11.03\% \\
\textbf{AIRFLOW} & $0.2524~(0.0005)$ & $0.3010~(0.0267)$ & 16.15\% \\
\textbf{ECG} & $0.2530~(0.0011)$ & $0.2185~(0.0351)$ & -13.64\% \\
\textbf{Mean of mean RMSE} & 0.1976 & 0.2185 & 9.57\% \\
\hline
\end{tabular}
\caption{Bayes-CATSI: Results for the test dataset for individual missing values showing RMSE mean and 95\% confidence interval for 30 independent experimental runs (CATSI) and  30  posterior samples (Bayes-CATSI). Mean RMSE Relative Improvement gives an idea of the performance improvement between CATSI and Bayes-CATSI per analyte where negative sign shows that Partial Bayes-CATSI performs worse than CATSI while positive sign shows that Bayes-CATSI performs better than CATSI.}
\label{tab:indMissing2}

\end{table*}

Tables \ref{tab:indMissing} and \ref{tab:indMissing2}  report the results (RMSE) using the test dataset for the partial Bayes-CATSI and Bayes-CATSI models for 30 independent experimental runs (CATSI) and  30 posterior samples (Bayes)\textcolor{black}{, the missing values in this case are of length one as described in the previous section}. We can observe that in both cases,  the RMSE standard deviation   for the Bayesian models is significantly lower than that of the original CATSI model. This observation suggests that the Bayesian models are more successful in reducing the uncertainty in predictions from   CATSI that features conventional  deterministic weights and biases, whereas \textcolor{black}{partial Bayes/}Bayes-CATSI represents them as probability distributions. {In Table \ref{tab:indMissing},  we find that Partial Bayes-CATSI underperforms CATSI for most of the analytes. We notice that out of 12 analytes, only 3 namely 'F4-M1', 'O2-M1' and 'CHEST' show a lower accuracy (higher RMSE mean) with Partial Bayes-CATSI, while the remaining 9 favor CATSI. Furthermore, CATSI demonstrates a 3.76\% better overall performance compared to Partial Bayes-CATSI, indicating that the addition of only a few Bayesian layers does not lead to meaningful performance gains. In Table \ref{tab:indMissing2}, we find that in contrast to the previous comparison, Bayes-CATSI consistently outperforms CATSI in 8 out of 12 analytes. The mean RMSE values for Bayes-CATSI are lower (better) across these analytes namely 'F3-M2','F4-M1','O2-M1','E1-M2','Chin1-Chin2','ABD','CHEST' and 'AIRFLOW'; with an overall improvement of 9.57\%. This highlights the significant benefit of incorporating Bayesian layers into every component (weight and biases) of the model. 

\begin{table}[htbp!]
\centering
\small
\caption{ Rate  of missing data (missing rates in \%) in consecutive missing value datasets compared with the ground-truth (actual)   original missing values as given in Table \ref{tab:data_table} datasets for the Partial Bayes-CATSI model, using 2000 samples.} 
\label{tab:consecutivepartialDataDesc}
\begin{tabular}{ c|c|c|c|c|c }
\hline
    & \textbf{Actual(\%)} & \multicolumn{4}{c }{\textbf{Consecutive Missing(\%)}} \\ \hline
    &      & \textbf{m=2}     & \textbf{m=3}     & \textbf{m=4}     & \textbf{m=5}     \\ \hline
\textbf{F3-M2}      & 0.165                         & 0.582            & 0.577            & 0.588            & 0.555            \\  
\textbf{F4-M1}      & 0.174                         & 0.563            & 0.578            & 0.599            & 0.600            \\  
\textbf{C3-M2}      & 0.164                         & 0.588            & 0.558            & 0.592            & 0.577            \\  
\textbf{C4-M1}      & 0.157                         & 0.569            & 0.569            & 0.601            & 0.550            \\  
\textbf{O1-M2}      & 0.167                         & 0.580            & 0.586            & 0.559            & 0.566            \\  
\textbf{O2-M1}      & 0.179                         & 0.576            & 0.601            & 0.583            & 0.585            \\ 
\textbf{E1-M2}      & 0.172                         & 0.575            & 0.593            & 0.553            & 0.581            \\  
\textbf{Chin1-Chin2}& 0.157                         & 0.561            & 0.580            & 0.547            & 0.538            \\  
\textbf{ABD}        & 0.154                         & 0.577            & 0.558            & 0.578            & 0.623            \\  
\textbf{CHEST}      & 0.158                         & 0.566            & 0.571            & 0.549            & 0.559            \\  
\textbf{AIRFLOW}    & 0.169                         & 0.576            & 0.584            & 0.564            & 0.586            \\  
\textbf{ECG}        & 0.165                         & 0.588            & 0.556            & 0.584            & 0.590            \\ \hline
\end{tabular}
\end{table}

\begin{table}[htbp!]
\centering
\small
\caption{Rate of missing data (missing rates in \%) in consecutive missing value datasets compared with the ground-truth(actual) original missing values for the Bayes-CATSI model, using 500 samples.}
\label{tab:consecutiveFullDataDesc}
\begin{tabular}{ c|c|c|c|c|c }
\hline
    & \textbf{ Actual (\%)} & \multicolumn{4}{c }{\textbf{Consecutive Missing(\%)}} \\ \hline
    &         & \textbf{m=2}     & \textbf{m=3}     & \textbf{m=4}     & \textbf{m=5}     \\ \hline
\textbf{F3-M2}      & 0.1583              & 0.5892           & 0.6308           & 0.5992           & 0.5667           \\ \hline
\textbf{F4-M1}      & 0.155               & 0.5867           & 0.6100           & 0.5883           & 0.5942           \\  
\textbf{C3-M2}      & 0.1725              & 0.5800           & 0.5950           & 0.5750           & 0.5542           \\ 
\textbf{C4-M1}      & 0.165               & 0.5800           & 0.6133           & 0.5208           & 0.5442           \\  
\textbf{O1-M2}      & 0.1717              & 0.6017           & 0.5200           & 0.5950           & 0.6292           \\  
\textbf{O2-M1}      & 0.145               & 0.5750           & 0.5483           & 0.6200           & 0.5933           \\  
\textbf{E1-M2}      & 0.1658              & 0.5675           & 0.5300           & 0.5933           & 0.5817           \\  
\textbf{Chin1-Chin2}& 0.1700              & 0.5350           & 0.5667           & 0.5567           & 0.5333           \\  
\textbf{ABD}        & 0.1625              & 0.6067           & 0.5667           & 0.5183           & 0.6283           \\  
\textbf{CHEST}      & 0.1583              & 0.5425           & 0.5592           & 0.5775           & 0.5433           \\  
\textbf{AIRFLOW}    & 0.1725              & 0.5583           & 0.6300           & 0.6208           & 0.4750           \\  
\textbf{ECG}        & 0.1817              & 0.5808           & 0.5492           & 0.5192           & 0.6400           \\ \hline
\end{tabular}
\end{table}

\begin{table*}[htbp]
\centering
\scriptsize
\caption{Results for consecutive missing value testing datasets: Partial Bayes-CATSI (Partial-BC) vs CATSI \textcolor{black}{with $m$ representing length of the generated consecutive missing values in the dataset}}
\label{tab:resultsPartialCons}
\begin{tabular}{|c|c|c|c|c|c|c|c|c|}
\hline
\textbf{} & \multicolumn{2}{c|}{\textbf{m=2}}    & \multicolumn{2}{c|}{\textbf{m=3}} & \multicolumn{2}{c|}{\textbf{m=4}} & \multicolumn{2}{c|}{\textbf{m=5}} \\
\textbf{} & \textbf{partialBC} & \textbf{CATSI} & \textbf{partialBC}    & \textbf{CATSI}    & \textbf{partialBC}    & \textbf{CATSI}    & \textbf{partialBC}    & \textbf{CATSI}    \\ \hline
\textbf{F3-M2}       & $0.1234~(0.0004)$ & $0.1261~(0.0097)$ & $0.1353~(0.0007)$ & $0.1339~(0.0122)$ & $0.1326~(0.0006)$ & $0.1394~(0.0132)$ & $0.1280~(0.0007)$ & $0.1321~(0.0135)$  \\ \hline
\textbf{F4-M1}       & $0.1572~(0.0008)$ & $0.1549~(0.0208)$ & $0.1495~(0.0009)$ & $0.1487~(0.0187)$ & $0.1563~(0.0005)$ & $0.1574~(0.0123)$ & $0.1428~(0.0007)$ & $0.1518~(0.0156)$ \\  \hline 
\textbf{C3-M2}       & $0.1293~(0.0006)$ & $0.1246~(0.0146)$ & $0.1173~(0.0006)$ & $0.1163~(0.0107)$ & $0.1347~(0.0009)$ & $0.1214~(0.0101)$ & $0.0984~(0.0006)$ & $0.1142~(0.0117)$ \\  \hline 
\textbf{C4-M1}       & $0.1493~(0.0005)$ & $0.1262~(0.0120)$ & $0.1233~(0.0003)$ & $0.1301~(0.0176)$ & $0.1311~(0.0006)$ & $0.1223~(0.0124)$ & $0.1606~(0.0005)$ & $0.1414~(0.0113)$  \\  \hline
\textbf{O1-M2}       & $0.1472~(0.0003)$ & $0.1526~(0.0150)$ & $0.1616~(0.0004)$ & $0.1476~(0.0129)$ & $0.1729~(0.0007)$ & $0.1499~(0.0127)$ & $0.1671~(0.0008)$ & $0.1463~(0.0120)$  \\  \hline
\textbf{O2-M1}       & $0.1443~(0.0004)$ & $0.1298~(0.0084)$ & $0.1476~(0.0007)$ & $0.1216~(0.0096)$ & $0.1366~(0.0005)$ & $0.1265~(0.0106)$ & $0.1247~(0.0006)$ & $0.1231~(0.0084)$  \\  \hline
\textbf{E1-M2}       & $0.1567~(0.0002)$ & $0.1620~(0.0157)$ & $0.1763~(0.0007)$ & $0.1672~(0.0172)$ & $0.1842~(0.0007)$ & $0.1791~(0.0130)$ & $0.1671~(0.0006)$ & $0.1694~(0.0132)$  \\  \hline
\textbf{Chin1-Chin2} & $0.1465~(0.0007)$ & $0.1143~(0.0156)$ & $0.1024~(0.0008)$ & $0.1137~(0.0137)$ & $0.1281~(0.0008)$ & $0.1126~(0.0136)$ & $0.0992~(0.0006)$ & $0.1092~(0.0128)$  \\   \hline
\textbf{ABD}         & $0.2659~(0.0003)$ & $0.2545~(0.0041)$ & $0.2543~(0.0003)$ & $0.2508~(0.0082)$ & $0.2553~(0.0003)$ & $0.2593~(0.0082)$ & $0.2334~(0.0003)$ & $0.2453~(0.0057)$  \\   \hline
\textbf{CHEST}       & $0.3009~(0.0005)$ & $0.2605~(0.0162)$ & $0.2405~(0.0005)$ & $0.2608~(0.0151)$ & $0.2362~(0.0003)$ & $0.2413~(0.0092)$ & $0.2628~(0.0004)$ & $0.2641~(0.0109)$  \\   \hline
\textbf{AIRFLOW}     & $0.2379~(0.0005)$ & $0.2348~(0.0193)$ & $0.2144~(0.0005)$ & $0.2328~(0.0162)$ & $0.2336~(0.0006)$ & $0.2370~(0.0162)$ & $0.2355~(0.0004)$ & $0.2473~(0.0222)$  \\   \hline
\textbf{ECG }        & $0.1610~(0.0010)$ & $0.1656~(0.0317)$ & $0.1774~(0.0016)$ & $0.1678~(0.0324)$ & $0.1412~(0.0010)$ & $0.1574~(0.0260)$ & $0.1657~(0.0014)$ & $0.1618~(0.0269)$ \\   \hline
\textbf{Mean of Mean RMSD }        & 0.1744 & 0.1522 & 0.1602 & 0.1619 & 0.1489 & 0.1606 & 0.1519 & 0.1495 \\ \hline 
\textbf{Relative Improvement }   & - & 12.73\% & 1.05\% & - & 7.29\% & - & 1.58\% & - \\ \hline 
\end{tabular}
\end{table*}

\begin{table*}[htbp]
\centering
\scriptsize
\caption{Results for consecutive missing value testing datasets: Bayes-CATSI(BC) vs CATSI \textcolor{black}{with $m$ representing length of the generated consecutive missing values in the dataset}}
\label{tab:resultsFullCons}
\begin{tabular}{|c|c|c|c|c|c|c|c|c|}
\hline
\textbf{} & \multicolumn{2}{c|}{\textbf{m=2}} & \multicolumn{2}{c|}{\textbf{m=3}} & \multicolumn{2}{c|}{\textbf{m=4}} & \multicolumn{2}{c|}{\textbf{m=5}} \\
\textbf{} & \textbf{BC}   & \textbf{CATSI}   & \textbf{BC}   & \textbf{CATSI}   & \textbf{BC}   & \textbf{CATSI}   & \textbf{BC}   & \textbf{CATSI}   \\ \hline
\textbf{F3-M2}       & $0.1658 ~(0.0005)$ & $0.1746 ~(0.0215)$ & $0.2021 ~(0.0008)$ & $0.1772 ~(0.0233)$ & $0.2156 ~(0.0010)$ & $0.1893 ~(0.0177)$ & $0.1881 ~(0.0007)$ & $0.2040 ~(0.0253)$ \\ \hline
\textbf{F4-M1}       & $0.2007 ~(0.0005)$ & $0.2095 ~(0.0285)$ & $0.2182 ~(0.0006)$ & $0.2193 ~(0.0346)$ & $0.1871 ~(0.0012)$ & $0.2091 ~(0.0251)$ & $0.2007 ~(0.0005)$ & $0.1993 ~(0.0219)$ \\ \hline
\textbf{C3-M2}       & $0.1459 ~(0.0007)$ & $0.1709 ~(0.0197)$ & $0.1518 ~(0.0007)$ & $0.1617 ~(0.0245)$ & $0.1395 ~(0.0006)$ & $0.1573 ~(0.0169)$ & $0.1623 ~(0.0008)$ & $0.1771 ~(0.0241)$ \\ \hline
\textbf{C4-M1}       & $0.1725 ~(0.0007)$ & $0.1672 ~(0.0305)$ & $0.2048 ~(0.0007)$ & $0.1586 ~(0.0188)$ & $0.1586 ~(0.0008)$ & $0.1676 ~(0.0197)$ & $0.1599 ~(0.0011)$ & $0.1693 ~(0.0292)$ \\ \hline
\textbf{O1-M2}       & $0.2614 ~(0.0012)$ & $0.2257 ~(0.0206)$ & $0.1881 ~(0.0010)$ & $0.2222 ~(0.0302)$ & $0.2089 ~(0.0007)$ & $0.2171 ~(0.0226)$ & $0.2267 ~(0.0010)$ & $0.2364 ~(0.0248)$ \\ \hline
\textbf{O2-M1}       & $0.2795 ~(0.0008)$ & $0.2026 ~(0.0216)$ & $0.2213 ~(0.0008)$ & $0.2299 ~(0.0364)$ & $0.2191 ~(0.0009)$ & $0.2014 ~(0.0262)$ & $0.1655 ~(0.0007)$ & $0.1680 ~(0.0178)$ \\ \hline
\textbf{E1-M2}       & $0.2406 ~(0.0005)$ & $0.2256 ~(0.0213)$ & $0.2931 ~(0.0009)$ & $0.2226 ~(0.0221)$ & $0.1902 ~(0.0007)$ & $0.2143 ~(0.0195)$ & $0.2068 ~(0.0004)$ & $0.2139 ~(0.0210)$ \\ \hline
\textbf{Chin1-Chin2} & $0.1415 ~(0.0008)$ & $0.1425 ~(0.0151)$ & $0.1880 ~(0.0009)$ & $0.1591 ~(0.0192)$ & $0.1593 ~(0.0010)$ & $0.1422 ~(0.0199)$ & $0.1592 ~(0.0007)$ & $0.1478 ~(0.0197)$ \\ \hline
\textbf{ABD}         & $0.3037 ~(0.0009)$ & $0.2752 ~(0.0166)$ & $0.2887 ~(0.0005)$ & $0.2994 ~(0.0175)$ & $0.2740 ~(0.0010)$ & $0.2611 ~(0.0180)$ & $0.2957 ~(0.0005)$ & $0.3016 ~(0.0204)$ \\ \hline
\textbf{CHEST}       & $0.2398 ~(0.0006)$ & $0.2955 ~(0.0225)$ & $0.3499 ~(0.0006)$ & $0.3074 ~(0.0214)$ & $0.3212 ~(0.0005)$ & $0.3076 ~(0.0194)$ & $0.2888 ~(0.0006)$ & $0.2668 ~(0.0260)$ \\ \hline
\textbf{AIRFLOW}     & $0.3409 ~(0.0012)$ & $0.2878 ~(0.0252)$ & $0.2987 ~(0.0008)$ & $0.2890 ~(0.0277)$ & $0.3358 ~(0.0011)$ & $0.2996 ~(0.0248)$ & $0.2604 ~(0.0007)$ & $0.2688 ~(0.0211)$ \\ \hline
\textbf{ECG}         & $0.2182 ~(0.0016)$ & $0.2084 ~(0.0244)$ & $0.2566 ~(0.0012)$ & $0.2089 ~(0.0353)$ & $0.2332 ~(0.0013)$ & $0.2022 ~(0.0310)$ & $0.2124 ~(0.0006)$ & $0.2013 ~(0.0420)$ \\ \hline
\textbf{Mean of Mean RMSD }   & 0.1942 & 0.1799 & 0.2067 & 0.2072 & 0.2233 & 0.2015 & 0.2037 & 0.2038 \\ \hline 
\textbf{Relative Improvement }   & - & 7.37\% & 0.24\% & - & - & 9.75\% & 0.05\% & - \\ \hline 

\end{tabular}
\end{table*}

\subsection{Consecutive Missingness}

In the field of medicine, individual missing values are rare \cite{Yin2020ContextAwareTS} and consecutive missing values present a more challenging and realistic scenario in clinical settings. We address this challenge by evaluating our model on consecutive missing values, reflecting the complexities of real-world medical data imputation. We vary the length  of the missing values from $m=2$ to $m=5$. Table \ref{tab:consecutivepartialDataDesc}  presents the missing rates for consecutive missing datasets, which can be  calculated as the number of missing time steps divided by the total time steps. The corresponding lengths of consecutive missing values are represented by $m$. We use this dataset for Partial Bayes-CATSI which contains 2000  samples \textcolor{black}{which corresponds to 10 seconds data at a frequency of 200 Hz}. Table \ref{tab:consecutiveFullDataDesc} describes the missing rates for the dataset used with Bayes-CATSI, which contains 500 samples \textcolor{black}{which corresponds to 2.5 seconds data at a frequency of 200 Hz}. We create  and add 5\% missing values to the pre-existing missing values in the original datasets. 

 Table \ref{tab:resultsPartialCons} and \ref{tab:resultsFullCons} show results of partial Bayes-CATSI  vs CATSI and Bayes-CATSI vs CATSI for consecutive missing value lengths $m$.
The results for Partial Bayes-CATSI   compared to CATSI, as shown in Table \ref{tab:resultsPartialCons}  demonstrate a better performance \textcolor{black}{compared to their individual missingness case.} For shorter lengths of consecutive missing values ($m=2$), partial Bayes-CATSI underperforms significantly, showing a 12.73\% worse RMSE compared to CATSI. This is a sharp decline when compared to the 3.76\% drop observed in individual missing value cases. However, as the missing sequence length increases, partial Bayes-CATSI  performance improves steadily. In the case of $m=3$, partial Bayes-CATSI  nearly matches CATSI at 1.05\% improvement and for the case of $m=4$, partial Bayes-CATSI  gets a peak improvement of 7.29\%. Despite this positive trend, partial Bayes-CATSI  performance declines again slightly for $m=5$, maintaining only a 1.58\% advantage over CATSI.  The results for Bayes-CATSI in Table \ref{tab:resultsFullCons} show more varied outcomes where in the case of  $m=2$ and $m=4$, Bayes-CATSI underperforms CATSI, \textcolor{black}{with CATSI performing 7.37\% and 9.75\% better than Bayes-CATSI}, respectively. On the other hand, Bayes-CATSI shows marginal improvement over CATSI for $m=3$ and $m=5$ by 0.24\% and 0.05\%, respectively. These results indicate that Bayes-CATSI, while effective in handling individual missing values, struggles to maintain performance with consecutive missing sequences which can be attributed to the limited data used for Bayes-CATSI, driven by computational constraints. As Bayes-CATSI samples more parameters, the model's complexity increases, making it more prone to multimodality and local minima issues, particularly when handling lesser number of samples. This trade-off between model complexity and lesser data samples may hinder its ability to fully capture the underlying patterns, especially with a smaller sample size, affecting its overall performance in comparison to non-probabilistic  models such as CATSI.

 In summary, both Partial Bayes-CATSI and Bayes-CATSI present lower standard deviations when compared to CATSI, suggesting more stable results. Partial Bayes-CATSI shows clear improvement as the length of consecutive missing values increases, whereas Bayes-CATSI performs better in individual missing value cases but struggles with consecutive missing value gaps. These observations point to the challenges in balancing model complexity and uncertainty quantification across different missing data patterns, with each approach having strengths in different scenarios.

\subsection{Confidence intervals}

We next visualise how the Bayes-CATSI model estimates missing features along with uncertainties (confidence interval) for selected patients in the dataset. 
 Figure \ref{fig:confint} illustrates the 'Chin1-Chin2' feature for a single patient, comparing the original dataset with missing values as shown in Panel A to the imputed dataset as shown in Panel B. It also presents the confidence intervals of the predictions generated by the Bayes-CATSI model. Figure \ref{fig:confint}-Panel B highlights the vertical lines that mark the positions of the 26 missing values in the time series, where it is evident that the 5\% and 95\% confidence intervals for the imputed values closely align with the imputed data.    We refer to Figure \ref{fig:dists} for enhanced visualisation of the model's predictions, which displays the distribution of predictions across 26 subplots—each corresponding to a missing value in the original dataset for the specified analyte. In Figure \ref{fig:dists}, the red lines indicate the 5\% and 95\% percentile values on either side of each subplot, providing a clear representation of the uncertainities projected in the predictions.

\begin{figure*}[htbp!]
  \centering
  \includegraphics[width=0.90\linewidth]{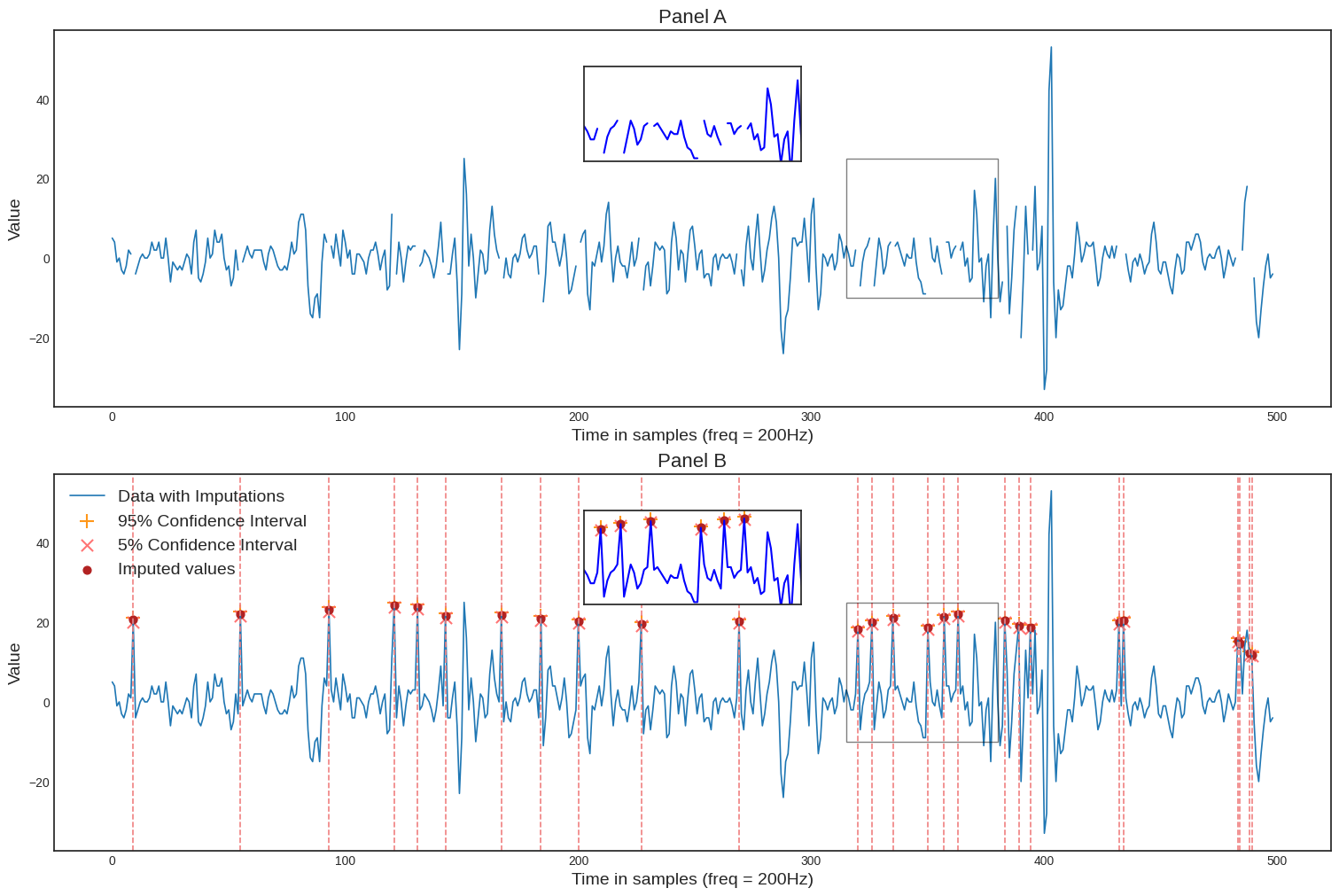}
  \caption{Visualisation of estimations (prediction and confidence interval) by Bayes-CATSI for imputing the missing values in the given patient sample from the dataset. Panel A corresponds to the original data with missing values, while Panel B corresponds to the data featuring  the imputations as predicted by Bayes-CATSI.}
  \label{fig:confint}
\end{figure*}

\begin{figure*}[htbp!]
  \centering
  \includegraphics[width=1.0\linewidth]{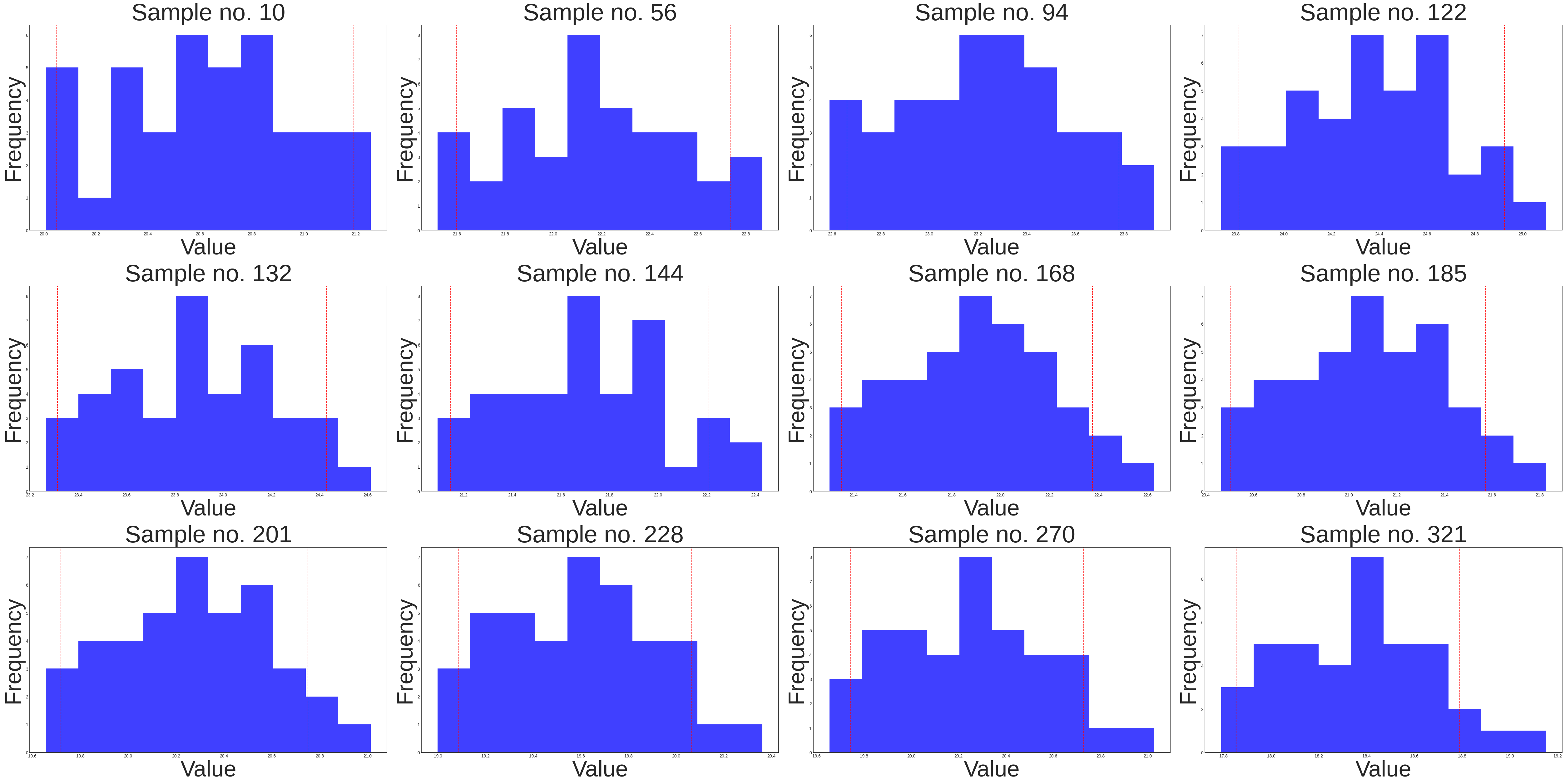}
  \captionof{figure}{Distribution visualization of imputations per missing sample as depicted in Figure \ref{fig:confint}. \textcolor{black}{The distributions of first 12 missing samples are shown. The red line on either side of each plot indicates the 5\% and 95\% percentile values.}}

  \label{fig:dists}
\end{figure*}

\section{Discussion}

We presented Bayes-CATSI that employs variational inference and integrates Bayesian deep learning layers into every component of the framework including context-aware recurrent imputation, cross-feature imputation,  and a fusion layer with a trainable weight that merges the results from these two components. We implemented and evaluated  the partial Bayes-CATSI model, which incorporated Bayesian model via posterior distribution into the CATSI framework. We also developed  the partial Bayes-CATSI model by adding Bayesian LSTM and Bayesian GRU layers and compared the results with the CATSI model.

The Bayesian layers integrated in our respective models offer significant advantages in handling uncertainty, a critical aspect of imputation in clinical datasets. Uncertainty quantification allows the model to express confidence in its predictions, particularly useful for clinicians who rely on these predictions for decision-making. The variational inference approach used in our framework ensures efficient and accurate training of the Bayesian layers, enabling reliable uncertainty estimates. Our framework builds upon prior work in deep learning for time series imputation, particularly the context-aware deep learning framework  by Yin et al. \cite{Yin2020ContextAwareTS} and the bidirectional RNN models for imputation  by Cao et al.\cite{NEURIPS2018_734e6bfc}. We extended these approaches by incorporating uncertainty quantification through variational inference\cite{pmlr-v37-blundell15} into the medical time series imputation process. This integration enabled  Bayes-CATSI to successfully quantify uncertainty while simultaneously achieving improved imputation predictions, demonstrating its effectiveness in handling both aspects.
 
 Despite these advancements, there are certain limitations that need to be addressed. One key limitation is that our current model has been tested on a relatively small number of samples due to computational constraints. Although the results  are promising, Bayesian model sampling strategies such as variational inference  require convergence  to sample the  posterior distribution for accurate predictions. Limited sample sizes may restrict the model’s ability to fully capture the underlying patterns and uncertainties, potentially impacting the accuracy of data imputation.  In the case of neural networks and deep learning models, it has been shown that convergence issues arise due to irregular and multimodal posterior distributions in neural networks \cite{chandra2024bayesian,wilson2020case}. In case the Bayesian model has not converged properly, we would get the accuracy that is far inference to conventional deep learning models. This can happen when the dataset is very noisy, inappropriate model architecture , and there is not enough data for Bayesian model convergence. Furthermore, we also note that the variational inference requires a prior distribution, which we assume to be a normal distribution. However, a major challenge in Bayesian deep learning has been the need to have an appropriate prior distribution \cite{fortuin2022priors} to effectively sample irregular and multimodal posterior distributions





Future studies can explore the results of Bayes-CATSI on larger datasets to fully evaluate its performance. Expanding the dataset size would allow for a more comprehensive assessment of the model's capabilities and provide further insights into its practical applicability in diverse clinical settings. Additionally, further research could investigate the integration of  as MCMC sampling \cite{hastings1970monte} into Bayes-CATSI framework. Although MCMC sampling is computationally intensive for larger datasets compared to variational inference \cite{pmlr-v37-blundell15}, it directly samples from the posterior distribution \cite{salimans2015markovchainmontecarlo}, whereas variational inference used in our study is an approximate approach.  Moreover, incorporating advanced deep learning models such as Transformer models\cite{10.5555/3295222.3295349} can potentially improve the performance of the context-aware recurrent imputation and cross-feature imputation components within Bayes-CATSI. Additionally, the Bayes-CATSI framework could be extended beyond medical time series, addressing imputation and prediction tasks in other domains such as environmental and climate data analysis.



\section{Conclusion}

In this paper, we presented and evaluated the Bayes-CATSI framework for the imputation of clinical time series data. The primary advantage of this model is its incorporation of uncertainty quantification into the CATSI model, leading to enhanced predictive performance.  Our results demonstrated that Bayes-CATSI outperforms CATSI by 9.57\% for individual missing values but partial Bayes-CATSI underperforms CATSI by 3.76\% for individual missing values. An analysis of four cases of consecutive missingness revealed that partial Bayes-CATSI has a better performance for consecutive missing values over CATSI model compared to Bayes-CATSI. These results demonstrated that incorporating Bayesian LSTM and Bayesian GRU layers effectively captures the complex temporal dynamics of the data using the uncertainty quantification provided by variational inference consequently leading to a better performance in the individual missingness case of Bayes-CATSI.

The Bayes-CATSI framework represents a significant advancement in the imputation of clinical time series data by effectively incorporating uncertainty quantification through Bayesian deep learning layers. The demonstrated improvements in predictive performance highlight the potential of this approach to provide more accurate and reliable imputations, ultimately aiding clinical decision-making processes.

\subsection*{Data and Code Availability}

Our data and code are available via GitHub repository \footnote{\url{https://github.com/pingala-institute/Bayes-medicaldataimputation}}.
 \bibliographystyle{ieeetr} 
 \bibliography{cas-refs}


\end{document}